\documentclass{article}

\PassOptionsToPackage{numbers}{natbib}
\usepackage[preprint]{neurips_2026}

\usepackage[utf8]{inputenc} 
\usepackage[T1]{fontenc}    
\usepackage{hyperref}       
\usepackage{url}            
\usepackage{booktabs}       
\usepackage{amsfonts}       
\usepackage{nicefrac}       
\usepackage{microtype}      
\usepackage{xcolor}         
\usepackage{natbib}

\usepackage{amsmath}
\usepackage{comment}
\usepackage{tabularx}
\usepackage{dirtytalk}
\usepackage{caption}
\usepackage{subcaption}
\usepackage{graphicx}
\usepackage{longtable}
\usepackage{lscape}
\usepackage{makecell}
\usepackage{multirow}
\usepackage{float}

\usepackage{soul}
\usepackage{xcolor}
\sethlcolor{yellow}

\title{Supervising Ralph Wiggum: Exploring a Metacognitive Co-Regulation Agentic AI Loop for Engineering Design}

\author{%
  Zeda Xu \\
  Department of Mechanical Engineering \\
  Carnegie Mellon University \\
  Pittsburgh, PA 15213 \\
  \texttt{zedaxu@cmu.edu} \\
  \And
  Nikolas Martelaro \\
  Human-Computer Interaction Institute\\
  Carnegie Mellon University \\
  Pittsburgh, PA, 15213 \\
  \texttt{nikmart@cmu.edu} \\
  \And
  Christopher McComb \thanks{Corresponding author} \\
  Department of Mechanical Engineering \\
  Carnegie Mellon University \\
  Pittsburgh, PA, 15213 \\
  \texttt{ccm@cmu.edu} \\
}

\begin{document}

\maketitle

\begin{abstract}

The engineering design research community has studied agentic AI systems that use Large Language Model (LLM) agents to automate the engineering design process. However, these systems are prone to some of the same pathologies that plague humans. Just as human designers, LLM design agents can fixate on existing paradigms and fail to explore alternatives when solving design challenges, potentially leading to suboptimal solutions. In this work, we propose (1) a novel Self-Regulation Loop (SRL), in which the Design Agent self-regulates and explicitly monitors its own metacognition, and (2) a novel Co-Regulation Design Agentic Loop (CRDAL), in which a Metacognitive Co-Regulation Agent assists the Design Agent in metacognition to mitigate design fixation, thereby improving system performance for engineering design tasks. In the battery pack design problem examined here, we found that the novel SRL and CRDAL systems generate designs with better performance, without significantly increasing the computational cost, compared to a plain Ralph Wiggum Loop (RWL) Further, the novel CRDAL generates designs with significantly better performance than SRL. Also, we found that the CRDAL system navigated through the latent design space more effectively than both SRL and RWL. The proposed system architectures and findings of this work provide practical implications for future development of agentic AI systems for engineering design. 

\end{abstract}

\section{Introduction}

Engineering design can be framed as a process that involves taking iterative design steps informed by design thinking and reasoning \cite{stahovich_learnit_1999, raina_learning_2019, rahman_predicting_2021, yuan_experimental_2018, brown_study_2024}. In an effort to assist or automate the design process, research in engineering design has widely explored the implementation of Artificial Intelligence (AI) \cite{allison_special_2022}. One of the key focus areas has been agent systems \cite{campbell_-design_1999, campbell_agent-based_2000}, especially reflective agents that solve design problems iteratively upon feedback \cite{moss_learning_2004, grecu_design_1996, mccomb_drawing_2016, mccomb_optimizing_2017}. These have been especially attractive for their ability to directly emulate the patterns of engineering teams and organizations. Recent Large Language Models have demonstrated incredible performance in reasoning, planning, and solving math and engineering problems \cite{brown_language_2020, kojima_large_2022, wei_chain--thought_2022, zoph_emergent_2022}. These support the emergence of modern LLM AI agents, which are LLM-powered automated agents capable of conducting a certain task independently without human intervention. These agents show promise for automated engineering design practices \cite{massoudi_agentic_2026, elrefaie_ai_2025, panta_meda_2025}. 

A variety of techniques have been proposed to supplement the capabilities of LLMs further. For instance, LLMs have been shown to improve task performance using feedback iteratively through self-reflection and self-refinement \cite{shinn_reflexion_2023, madaan_self-refine_2023}. In software engineering, practitioners have pioneered the Ralph Wiggum Loop\footnote{Named after the \textit{Simpsons} character Ralph Wiggum for a similar behavior pattern.}, in which an AI agent(s) runs continuously in a loop and repeatedly attempts the given task until success, to empower LLM self-reflection and more effectively implement LLM AI agents \cite{huntley_everything_2026}. In a Ralph Wiggum Loop, the success of the solution is determined by external validation rather than the agent's own judgment. The LLM AI agent(s) take the external feedback, inspect the output, reflect, and try to generate a better solution. This pattern resembles precursor reflective agentic systems from prior research \cite{moss_learning_2004, grecu_design_1996, mccomb_drawing_2016, mccomb_optimizing_2017}. The reflective process is also analogically similar to how designers are theorized to work, by reflection in and on their action \cite{schon_designing_1992, schon_reflective_2017, roozenburg_describing_1998, valkenburg_reflective_1998}. 


However, poorly regulated reflection can lend itself to design fixation. Specifically, design fixation occurs when designers adhere prematurely to limited paradigms and fail to see alternative solutions \cite{sio_fixation_2015, salvi_how_2024}. Design fixation and insufficient design exploration reduce design creativity and potentially design quality \cite{jansson_design_1991, linsey_study_2010, viswanathan_design_2013}. Much like human designers, AI systems may also exhibit design fixation \cite{chen_understanding_2025}, potentially leading to similar issues in creativity and design quality for AI design systems.

Mitigating fixation is challenging, but may be possible through appropriate metacognitive strategies. Metacognition is the cognition and monitoring of the cognitive process \cite{brown_knowing_1978, flavell_metacognition_1979}, and described by Kellogg as the "cognition about cognition, or thinking about thinking" \cite{kellogg_cognitive_2002, dixon_use_2012}. To mitigate design fixation and insufficient design exploration for human designers, research in cognitive science and human-computer interaction (HCI) has explored metacognitive support in the design process \cite{gmeiner_exploring_2023, gmeiner_exploring_2025}, whereby systems prompt users with questions and interactions to help scaffold their thinking and problem-solving process. 

Researchers in engineering design education have investigated Self-Regulated Learning (SRL) as a metacognitive self-regulation strategy for student learning \cite{zheng_profiling_2020, lawanto_students_2010, lawanto_self-regulated_2014}. Self-Regulated Learning emphasizes self-regulation, and prompts self-driven goal-setting, progress monitoring, and reflection to guide and support the learning process \cite{zimmerman_social_1989, boekaerts_self-regulated_1997, winne_studying_1998, pintrich_conceptual_2004}. A close variant of Self-Regulated Learning, Co-Regulated Learning, introduces metacognition through co-regulation with others as a different style of learning regulation \cite{mccaslin_self-regulated_2001, lim_co-regulation_2020, allal_assessment_2020}. Self-Regulated Learning and co-regulated learning have shown their effectiveness in improving student learning outcomes \cite{jansen_self-regulated_2019, yan_effects_2022, xu_meta-analysis_2023, sola-guirado_enhancing_2024}. Since designing can also be modeled as a learning process \cite{beguin_design_2003}, the success of self-regulation and co-regulation as metacognitive support in learning might be transferable to human designers in engineering design. 

As LLMs demonstrate strong capabilities in learning and reasoning \cite{brown_language_2020, kojima_large_2022, wei_chain--thought_2022, zoph_emergent_2022}, research has explored modeling metacognition in LLM agents, with substantial performance benefits \cite{wang_metacognitive_2024, renze_self-reflection_2024, zhou_metagent-p_2025, liu_position_2025}. Therefore, self-regulation and co-regulation as metacognitive support might also be beneficial to LLM agents in engineering design tasks. 

So far, there have been few attempts in the research community to use metacognitive support (through either self-regulation or co-regulation) to assist LLM agents in agentic AI systems for engineering design. To address this gap, we draw inspiration from metacognitive support for human designers and Self-Regulated/Co-Regulated Learning in engineering design education, and apply a separate metacognitive support agent to reflective agentic systems. In this paper, we propose (1) a novel Self-Regulation Loop (SRL), in which the Design Agent self-regulates and explicitly monitors its own metacognition, and (2) a novel Co-Regulation Design Agentic Loop (CRDAL), in which a Metacognitive Co-Regulation Agent assists the Design Agent in metacognition. We believe that such novel agentic AI system architectures can mitigate design fixation, leading to improved system performance in engineering design tasks. 

This leads to our research question: Do metacognitive self- and co-regulation processes improve agentic system performance for engineering design tasks? We hypothesize that (1) a metacognitive \textit{Self-Regulation} Loop improves system performance over a plain Ralph Wiggum Loop, (2) a metacognitive \textit{Co-Regulation} Design Agentic Loop improves system performance over a plain Ralph Wiggum Loop, and (3) both the metacognitive self-regulation and co-regulation agentic loops will mitigate design fixation and allow the agentic systems to explore a design space more effectively. 

To test our hypotheses, we compare the Self-Regulation Loop (SRL), and the Co-Regulation Design Agentic Loop (CRDAL), against a plain Ralph Wiggum Loop (RWL), through a battery pack cell configuration design problem. The agentic design systems are asked to design a battery pack that maximizes capacity while satisfying physical, thermal, and electrical constraints, using 18650 Lithium-ion battery cells. 

Our results confirm hypotheses (1) and (2) fully and also partially confirm hypothesis (3). In the design problem demonstrated in this paper, we found that our novel SRL and CRDAL systems generate designs with better performance, without significantly increasing the computational cost, compared to a plain Ralph Wiggum Loop (RWL). Also, we found that the CRDAL system navigated through the latent design space more effectively. This paper makes three primary contributions:
\begin{itemize}
    \item We examine three different agentic AI system architectures for engineering design tasks, including a novel Self-Regulation Loop (SRL) and Co-Regulation Design Agentic Loop (CRDAL) for metacognition-enabled agentic AI systems. The proposed system architectures provide practical implications for future development of agentic AI systems for engineering design. 
    
    \item We present early evidence on the effectiveness of the novel SRL and CRDAL systems for improved design solution performance and design space exploration. We also demonstrate the effectiveness of the other system, RWL, in completing an engineering design task. 

    \item We introduce a multi-disciplinary design problem and its related design evaluation, which differentiates agentic AI systems' performance. This can serve as a benchmark for agentic system performance in engineering design tasks. 
\end{itemize}

\section{Methodology}

To evaluate the effect of metacognitive co-regulation, we define a constrained battery pack design task, instantiate three agentic system architectures, and compare them under a common experimental protocol.

\subsection{Design Problem}

\begin{figure}[H]
    \centering
    \includegraphics[width=0.4\linewidth]{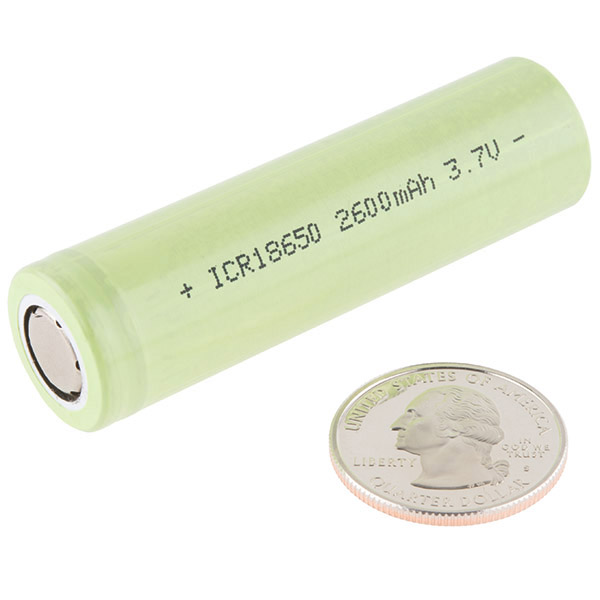}
    \caption{A Common 18650 Lithium Ion Battery.}
    \footnotemark
    \label{fig:18650_cell}
\end{figure}

\begin{figure}[H]
    \centering
    \includegraphics[width=0.6\linewidth]{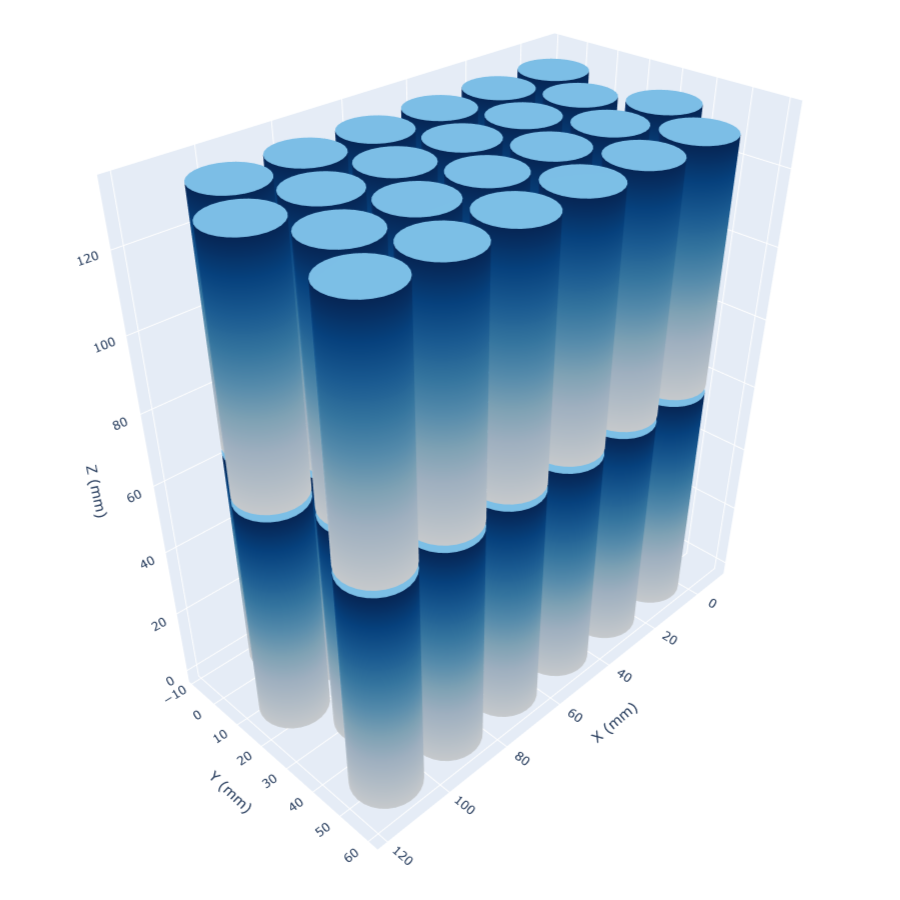}
    \caption{Example of a simple $6(\text{W}) \times 4(\text{D}) \times 2(\text{H}) $ battery pack composed of 18650 Lithium-ion battery cells using hexagonal close-packing. }
    \footnotemark
    \label{fig:battery_pack_example}
\end{figure}

We compare the agentic systems using a battery pack cell configuration design problem. This entails designing a battery pack that maximizes capacity while satisfying physical, thermal, and electrical constraints, using only 18650 Lithium-ion battery cells (Figure~\ref{fig:18650_cell}). In the design problem, the 18650 battery cells are hexagonally close-packed in a gridded pack, as illustrated in Figure~\ref{fig:battery_pack_example}. This task is a multi-disciplinary engineering design problem with multi-step design optimization and objective design performance evaluation. With multiple constraints and several distinct physics types, the design problem poses a significant challenge for the agentic systems examined here. 

\footnotetext{"Polymer Lithium Ion Battery - 18650 Cell (2600mAh, Solder Tab)" by SparkFunElectronics is licensed under CC BY 2.0. To view a copy of this license, visit https://creativecommons.org/licenses/by/2.0/?ref=openverse.} 

\subsubsection{Design Objective, Constraints, and Assumptions}

The agentic design systems are instructed to generate a battery pack design using only 18650 cells to satisfy all constraints, while maximizing capacity:
\begin{quote}
Design a $400 V$ battery pack with a minimum capacity of $50 Ah$, capable of continuously supplying at least $48 A$ of current draw while staying at or below $60 ^{\circ} C$ during operation, within a $1000mm \times 1000mm \times 250mm$ envelope.
\end{quote}

The 18650 cells have a diameter of 18 mm and a height of 65 mm. They are assumed to have a nominal voltage of 3.7 V, a nominal capacity of 2.5 Ah, and an internal resistance of 0.05 Ohm. The cells are assumed to be placed in an upright orientation (cylindrical axis vertical), in a grid pattern, using hexagonal close packing, and with uniform spacing between cells. Hexagonal close packing is a common and optimal packing method for compact battery pack designs using cylindrical Lithium-ion battery cells, such as 18650 cells \cite{pegel_pareto-optimal_2025, balasubramanian_study_2025}. The minimum spacing between cells is 2mm to allow for cooling and manufacturing tolerances. We also assume an ambient temperature of 20 degrees Celsius and passive cooling for the battery pack.

\subsubsection{Design Actions}

The design agent can generate designs using the following specific design actions: 

\begin{enumerate}
\item Defining cell locations: 

$\mathrm{CELL\_LOCATIONS}: [[x_1, y_1, z_1], [x_2, y_2, z_2], ...]$

Each $x, y, z$ triplet is the coordinates of a cell's location. This design action allows the agent to add or remove cells, and adjust the cells' positions within the battery pack. Adding or removing cells may affect one or more aspects of the battery pack's electrical performance (e.g., voltage, capacity, maximum current), its physical dimensions, and thermal performance. 

\item Defining cell connections: 

$\mathrm{CELL\_CONNECTIONS}: [N_{\mathrm{series}}, N_{\mathrm{parallel}}]$

$N_{\mathrm{series}}$ and $N_{\mathrm{parallel}}$ are the number of series and parallel connections in the battery pack. This design action allows the agent to define the number of series and parallel connections in the battery pack. Adjusting the number of series and parallel connections will change the electrical performance of the battery pack (e.g., voltage, capacity, maximum current). 

\item Defining cell spacing: 

$\mathrm{CELL\_SPACING}: [D_{\mathrm{spacing}}]$

$D_{\mathrm{spacing}}$ is the uniform spacing between battery cells in millimeters. This design action allows the agent to define the spacing between cells in a hexagonal close-packed arrangement. Adding or reducing the cell spacing will change the battery pack's physical dimensions and thermal performance. 

\end{enumerate}

\subsubsection{Design Evaluation and Assumptions}

The designs generated by the design agent are assessed through two approaches.

\begin{itemize}

\item A numerical \textit{evaluator} evaluates the design for mechanical performance (e.g., battery pack dimensions and design volume), thermal performance (e.g., maximum temperature of the cells under load), and electrical performance (e.g., voltage, capacity, maximum current). 

\item A numerical \textit{validator} checks for design validity, including physical validity (e.g., whether there are overlapping cells or insufficient spacing between cells) and electrical connection validity (e.g., whether the claimed connections are feasible with the claimed cell numbers). It also checks whether the design constraints (e.g., physical dimension constraints, electrical property constraints, cell temperature constraints) are met. 
\end{itemize}

Together, the numerical evaluator and the numerical validator objectively and comprehensively assess the performance of the generated battery pack designs.

\subsection{Agentic Design Systems}

We introduce the architecture and design of each agentic design system in the following paragraphs. \footnote{The source code and LLM prompts of the agentic systems are available at (https://github.com/cmudrc/Co\_Regulation\_Design\_Agent\_Loop\_IDETC). }

\subsubsection{Ralph Wiggum Loop (RWL)}

\begin{figure}[h]
    \centering
    \includegraphics[width=0.5\linewidth]{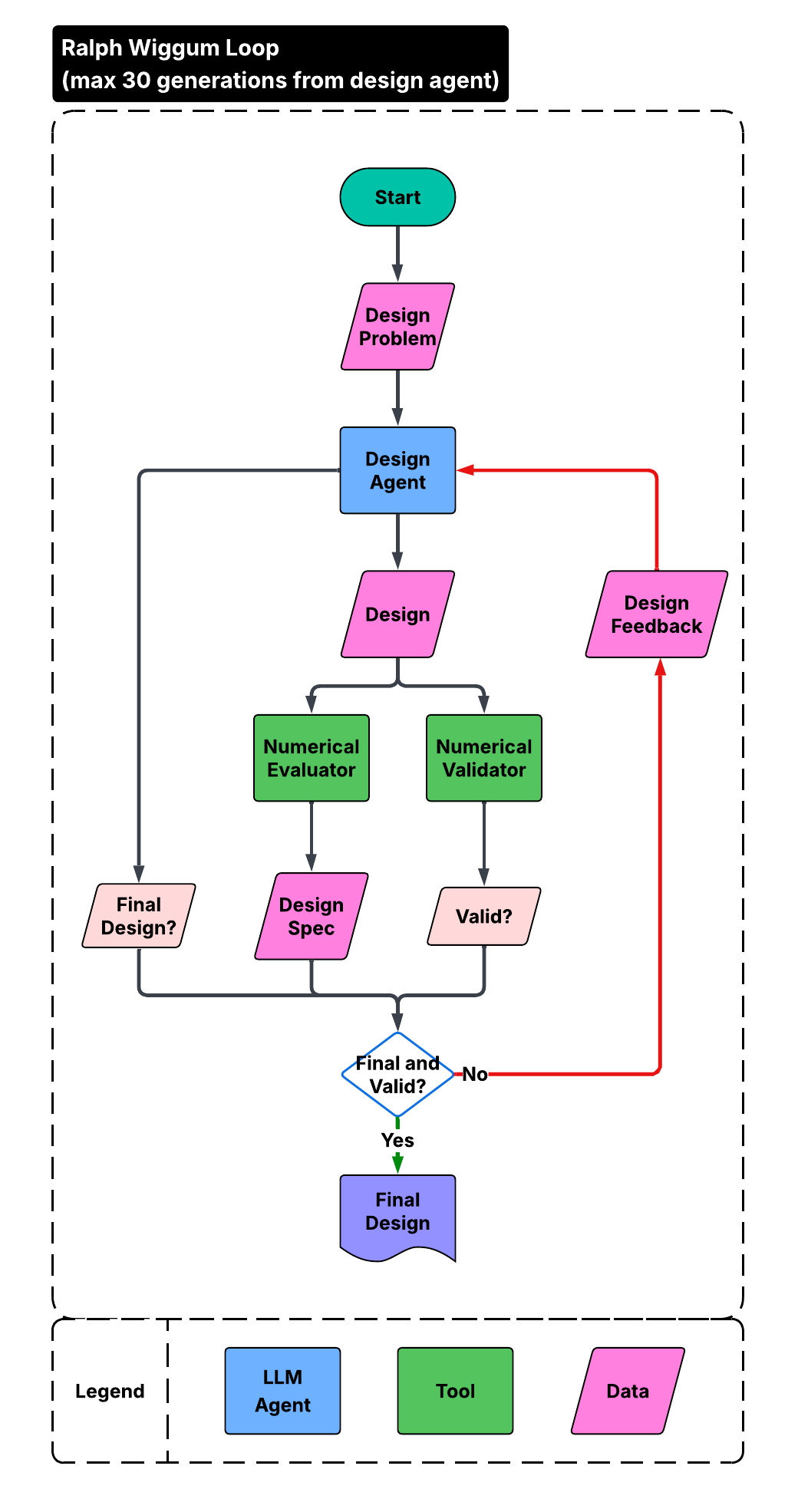}
    \caption{System flowchart of the Ralph Wiggum Loop (RWL). }
    \label{fig:RWL_system}
\end{figure}

In the RWL, as shown in Figure~\ref{fig:RWL_system}, the Design Agent is made to generate designs until a valid final design is given. After each design generation, the design solution is evaluated and validated by the numerical evaluator and validator. If the design is not valid, the design specification and validation information for the current design are provided to the Design Agent as Design Feedback (e.g., design attempt failed because of failing thermal constraints, the current design capacity and voltage, the current design dimensions and maximum cell temperature). The design agent uses the feedback, reflects on it, and then creates another design. Across multiple iterations, the Design Agent also has access to the full design history and previous Design Feedback.

Additionally, the Design Agent self-evaluates after each design generation. It can make another attempt to make further improvements, even when the design solution is valid. In this case, the Design Agent will also receive Design Feedback, and can reflect and generate another design. The Design Agent is informed that it may receive multiple rounds of feedback with valid designs. It is instructed to judge whether the design can be improved further, and to terminate only when it is confident that the design cannot be meaningfully improved further.

A maximum of 30 design generations is allowed for a given design problem. This value was determined during pilot testing, where we observed that the systems generated final designs in 30 design steps or fewer in most cases. 

The Design Agent can only terminate the design loop and output a \textit{final design solution} when two conditions are met: (1) it has received feedback from the numerical evaluator and validator, and the design is both valid and feasible; and (2) it is confident that the design cannot be meaningfully improved further.

\subsubsection{Self-Regulation Loop (SRL)}

\begin{figure}[h]
    \centering
    \includegraphics[width=0.6\linewidth]{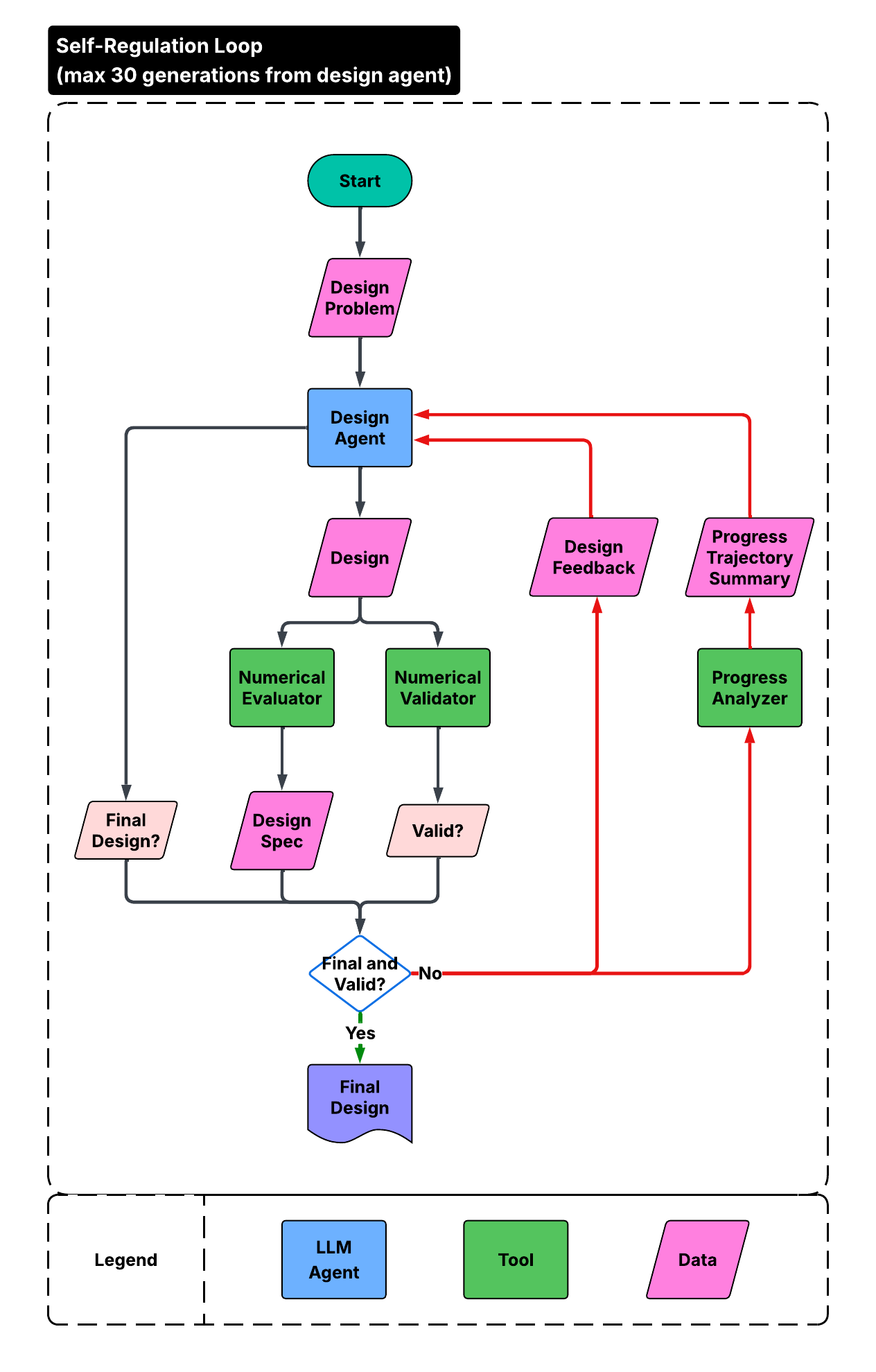}
    \caption{System flowchart of the Self-Regulation Loop (SRL). }
    \label{fig:SRL_system}
\end{figure}

The system architecture of the SRL is built upon the RWL, as shown in Figure~\ref{fig:SRL_system}. The difference lies in the feedback the Design Agent receives after each iteration. 

In SRL, when the Design Agent gets another attempt (i.e., if the design is not valid or it wants to make further improvements), a Progress Analyzer first takes the design step history, and explicitly shows the design progress trajectory and the trend summary (e.g., changes in capacity, which of the past attempts are valid designs) to the Design Agent, along with the Design Feedback. 

In addition, in SRL, the Design Agent is specifically instructed to set design goals, make plans, monitor its progress, try potential alternatives, and pursue higher performance designs. Also, for each design step, it is asked to assess its design progress from design history (improving, stalling, or regressing), identify the bottleneck metric or constraint, and think for design strategy for the next design iteration. In the Ralph loop, the Design Agent also implicitly performs goal-setting, planning, and progress monitoring when self-reflecting, whereas in the SRL, we made it very explicit that the agent should do so. The SRL has the same termination rule as the RWL, and is also allowed a maximum of 30 design generations for a given design problem.  

\subsubsection{Co-Regulation Design Agentic Loop (CRDAL)}

\begin{figure}[h]
    \centering
    \includegraphics[width=0.6\linewidth]{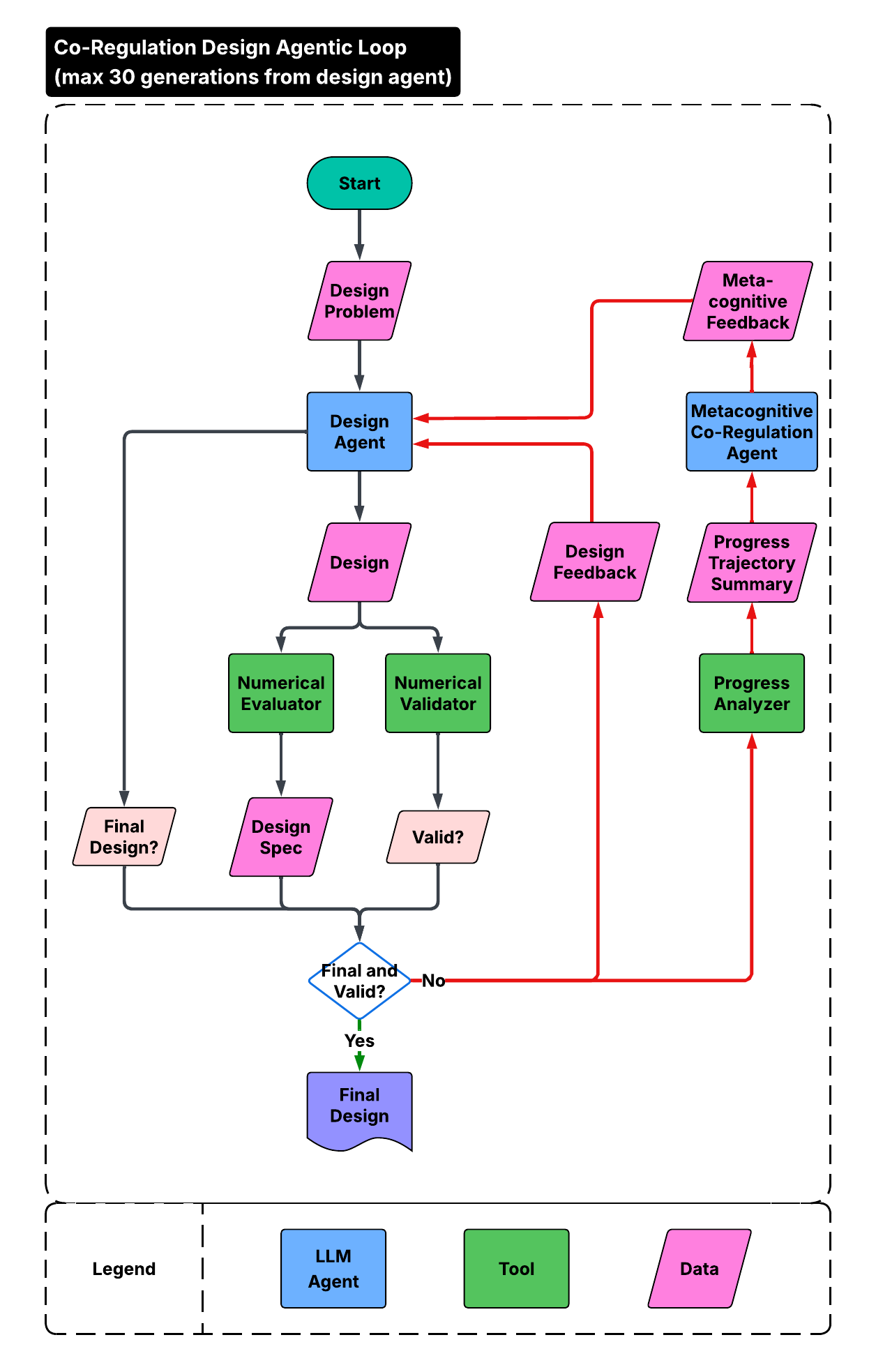}
    \caption{System flowchart of the Co-Regulation Design Agentic Loop (CRDAL). }
    \label{fig:CRDAL_system}
\end{figure}

Built upon SRL, the CRDAL adds a separate Metacognitive Co-Regulation Agent to the system, illustrated in Figure~\ref{fig:CRDAL_system}. The Metacognitive Co-Regulation Agent is a separate LLM agent from the Design Agent. In CRDAL, when the Design Agent gets another attempt (i.e., if the design is not valid or it wants to make further improvements), the Progress Analyzer takes the design step history, and explicitly shows the design progress trajectory and the trend summary to the Metacognitive Co-Regulation Agent. 

The Metacognitive Co-Regulation Agent then processes the Progress Trajectory Summary, analyzes the current design and the design history, and provides strategic Metacognitive Feedback to the Design Agent along with Design Feedback. Similar to the self-assessment in SRL, the Metacognitive Feedback includes an assessment of the Design Agent's progress (improving, stalling, or regressing), identification of the bottleneck metric or constraint, and a strategic suggestion for the next design iteration. In CRDAL, the Metacognitive Co-Regulation Agent is "supervising" the Design Agent. It provides design review and helps the "designer" think through a design problem, like a supervisor or a colleague. The CRDAL has the same termination rule as the previous two systems, and is also allowed a maximum of 30 design attempts for a given design problem.

\subsubsection{LLM Model Configuration}
All of the LLM agents in this work, including the Design Agent and the Metacognitive Co-Regulation Agent, are powered by Google DeepMind's
Gemini 3.1 Pro model. According to Google DeepMind's own benchmarking, at the time of writing, it is one of the most advanced and powerful LLMs for reasoning and solving complex math problems \cite{noauthor_gemini_2026}. The LLM model is allowed the highest reasoning budget for optimal performance.\footnote{Specifically, we used \texttt{gemini-3.1-pro-preview} model, Feb 2026 release. The model "thinking\_level" was set to "high" and the "temperature" was set to 1.0 for optimal performance, as recommended by the model developer \cite{noauthor_gemini_nodate}. The model details can be found on the developer website https://deepmind.google/models/model-cards/gemini-3-1-pro/. }

\subsection{Measurement}

For consistency and generalizability, all three systems were instructed to solve the same design problem 30 times\footnote{30 system runs for each system. This differs from the 30 maximum generation attempts the design agents have per system run. }. The agentic systems have no memory of previous or other systems' runs. For each system run, the LLM models are initialized with a different random seed, determined by a set base seed. The base seed is the same for all three systems. For each system run, the performance of the system is determined by the \textit{final designs} it outputs. If a system fails to provide a valid final design within the allowed 30 design steps, it is considered a failed design generation and excluded from the design performance comparison. 

For this work, our primary measure of system performance is the capacity of the generated battery packs. This is the objective that the agentic design systems are explicitly instructed to maximize. To better understand system behavior with respect to design fixation, we also examine their design trajectories and the location of final designs in the latent design space. 

We are also interested in the computational cost of the systems, as this often defines a sharp trade-off against system performance. In engineering design practice for complex systems, the primary computational bottleneck tends to be design evaluation, which often makes use of long-running simulations. We assume that the computational cost of such simulations will be larger than the computational costs associated with LLMs for any sufficiently complex task. Therefore, in this work, we focus on the number of steps taken before a final design is given as an indication of computational cost for a more meaningful comparison.

\section{Results}

We report results on three aspects of system behavior in the battery pack design problem: final design performance, computational cost, and design-space exploration.

\subsection{System Performance}

\begin{figure}[H]
    \centering
    \includegraphics[width=0.6\linewidth]{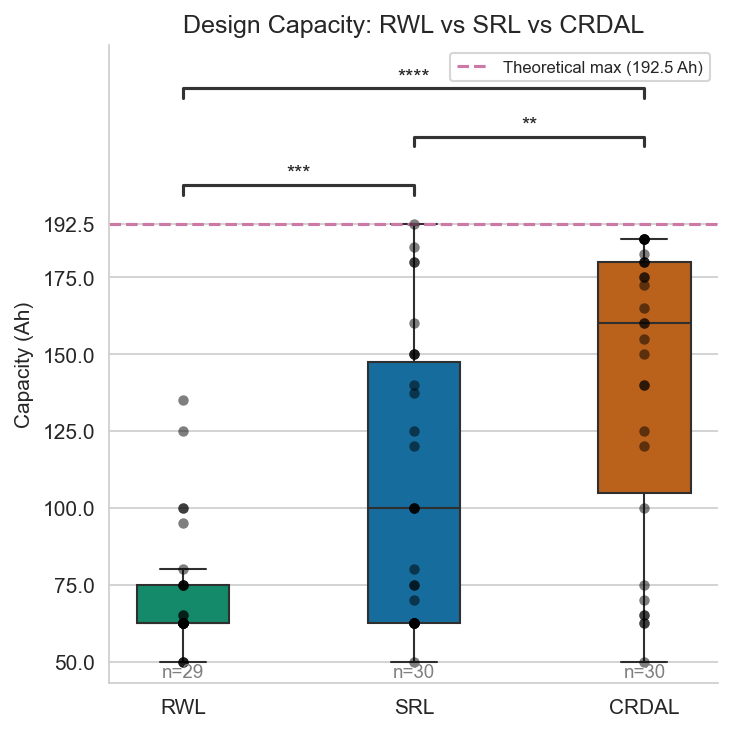}
    \caption{Design capacity of the battery pack created by each agentic design system. 
    p-value annotation legend:
    ns: $1.70\times10^{-2} < p$; 
    \quad $*: 1.00\times10^{-2} < p <= 1.70\times10^{-2}$; 
    \quad $**: 1.00\times10^{-3} < p <= 1.00\times10^{-2}$; 
    \quad $***: 1.00\times10^{-4} < p <= 1.00\times10^{-3}$; 
    \quad $****: p <= 1.00\times10^{-4}.$ }
    \label{fig:box_plot_RWL_SRL_CRDAL}
\end{figure}

The capacities of the final battery pack designs created by each agentic design system are shown in Figure~\ref{fig:box_plot_RWL_SRL_CRDAL}.  The dots in the figure show the capacities of the final battery pack designs for each run. The box plot shows the mean, 25\%, and 75\% quartiles of the valid final designs from the 30 runs. 

The success rate, design capacity range, mean, standard deviation, and median out of the 30 runs are shown in Table \ref{tab:summary_stat_capacity}. 

\begin{table}[h]
\caption{Summary statistics on the design capacity of battery pack designs created by agentic design systems. }

    \centering
    
    \begin{tabularx}{0.85\textwidth} { 
    | >{\centering\arraybackslash}X
    | >{\centering\arraybackslash}X
    | >{\centering\arraybackslash}p{0.15\textwidth} 
    | >{\centering\arraybackslash}p{0.1\textwidth} 
    | >{\centering\arraybackslash}p{0.1\textwidth} 
    | >{\centering\arraybackslash}p{0.1\textwidth}  | }
    
    \hline
    \textbf{System} & \textbf{Success Rate} & \textbf{Range \newline ($Ah$)} & \textbf{Mean \newline ($Ah$)} & \textbf{Std. \newline ($Ah$)} & \textbf{Median ($Ah$)} \\
    \hline 
    RWL & $29/30$ & $[50.00, 135.00]$ & $72.07$ & $20.52$ & $62.50$ \\ 
    \hline
    SRL & $30/30$ & $[50.00, 192.50]$ & $106.08$ & $45.60$ & $100.00$ \\ 
    \hline
    CRDAL & $30/30$ & $[50.00, 187.50]$ & $141.17$ & $48.45$ & $160.00$ \\ 
    \hline
    
\end{tabularx}
\label{tab:summary_stat_capacity}
\end{table}

Overall, all three systems are effective in completing the design task, generating valid final designs that satisfy design requirements and constraints in nearly all 30 runs, within the maximum allowed 30 design steps in each run. The RWL experienced a single run failure in which no valid final design was generated within 30 design steps. That run is excluded from further comparison. The plain Ralph Wiggum Loop (RWL) has the lowest design capacity range due to a lower upper bound, among the three systems. It also has lower average and median capacities than the other two systems. 

The SRL has the highest capacity in all attempts from all three systems at $192.5 Ah$. The CRDAL has a higher average and median capacity compared to RWL and SRL. It is also worth noting that RWL has a lower standard deviation in design capacity, suggesting higher consistency across runs. 

To answer our research question and test our hypotheses, we conducted statistical tests to unveil any differences in system performance. Specifically, we conducted ANOVA tests on the battery pack capacity for designs generated by each agentic design system. There are significant differences in design capacity across systems: $F(2, 86) = 21.606$, $p < 0.001$, $\eta _p^2 = 0.334$. The results indicate that different systems generated battery pack designs with significantly different capacities. Further, we performed follow-up pairwise $t$-tests between systems, with a Bonferroni correction adjusted alpha value of $\alpha = 0.05 / 3 = 0.017$. The results are shown in Table \ref{tab:t-test_capacity}.

There are significant differences in design capacity in pairwise comparisons between designs generated by the RWL system and the SRL system, between those generated by the RWL system and the CRDAL system, and between those generated by the SRL system and the CRDAL system. 

\begin{table}[h]
\caption{T-test results on the design capacity in pairwise comparisons between agentic design systems. }
    \centering
    
    \begin{tabularx}{0.85\textwidth} { 
    | >{\centering\arraybackslash}p{0.25\textwidth} 
    | >{\centering\arraybackslash}X | }
    
    \hline
    \textbf{Comparison} \textbf{Group} & \textbf{T-test results} \\
    \hline 
    RWL vs SRL & $t(40.6) = 3.715, p = 0.001,$ Cohen's $d = 0.956$ \\ 
    \hline
    RWL vs CRDAL & $t(39.4) = 7.174, p < 0.001,$ Cohen's $d = 1.846$ \\
    \hline
    SRL vs CRDAL & $t(57.8) = 2.888, p = 0.005,$ Cohen's $d = 0.746$ \\
    \hline
    
\end{tabularx}
\label{tab:t-test_capacity}
\end{table}

\subsection{Computational Cost}

\begin{figure}[H]
    \centering
    \includegraphics[width=0.6\linewidth]{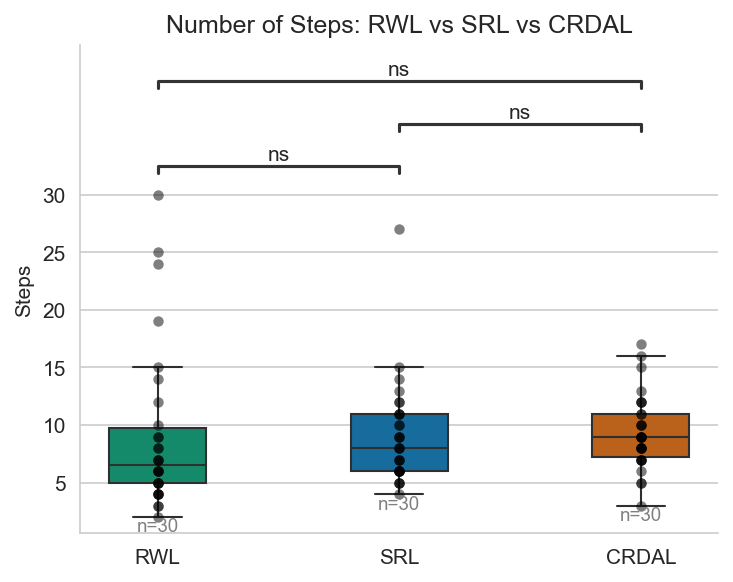}
    \caption{Number of design steps taken before final design by each agentic design system. 
    p-value annotation legend:
    ns: $1.70\times10^{-2} < p$; 
    \quad $*: 1.00\times10^{-2} < p <= 1.70\times10^{-2}$; 
    \quad $**: 1.00\times10^{-3} < p <= 1.00\times10^{-2}$; 
    \quad $***: 1.00\times10^{-4} < p <= 1.00\times10^{-3}$; 
    \quad $****: p <= 1.00\times10^{-4}.$ }
    \label{fig:box_plot_RWL_SRL_CRDAL_steps}
\end{figure}

The number of design steps taken by each agentic design system is shown in Figure~\ref{fig:box_plot_RWL_SRL_CRDAL_steps}. The dots in the figure show the steps taken for each run. The box plot shows the mean, 25\%, and 75\% quartiles of steps taken from the 30 runs. The range, mean, standard deviation, and median of the steps taken before final design out of the 30 runs are shown in Table \ref{tab:summary_stat_steps}.

\begin{table}[h]
\caption{Summary statistics on the number of steps taken before final design by agentic design systems. }

    \centering
    
    \begin{tabularx}{0.95\textwidth} { 
    | >{\centering\arraybackslash}X
    | >{\centering\arraybackslash}X
    | >{\centering\arraybackslash}X
    | >{\centering\arraybackslash}X
    | >{\centering\arraybackslash}X | }
    
    \hline
    \textbf{System}  & \textbf{Range (steps)} & \textbf{Mean (steps)} & \textbf{Std. (steps)} & \textbf{Median (steps)} \\
    \hline 
    RWL & $[2, 30]$ & $9.07$ & $7.02$ & $6.50$ \\ 
    \hline
    SRL & $[4, 27]$ & $9.10$ & $4.45$ & $8.00$ \\ 
    \hline
    CRDAL & $[3, 17]$ & $9.37$ & $3.20$ & $9.00$ \\ 
    \hline
    
\end{tabularx}
\label{tab:summary_stat_steps}
\end{table}

We conducted ANOVA tests on the number of steps taken before a final design is given for each agentic design system. There are no significant differences in the number of steps taken across systems: $F(2, 87) = 0.031 $, $p = 0.97$, $\eta _p^2 = 0.001$. The results indicate that different systems did not take significantly different numbers of steps before giving a final design solution. 

Further, we performed follow-up pairwise $t$-tests between systems, with a Bonferroni correction adjusted alpha value of $\alpha = 0.05 / 3 = 0.017$. The results are shown in Table \ref{tab:t-test_step}. There are no significant differences found in the number of design steps taken before the final design in pairwise comparisons between the systems. 

\begin{table}[h]
\caption{T-test results on the number of steps taken before final design in pairwise comparisons between agentic design systems. }

    \centering
    
    \begin{tabularx}{0.85\textwidth} { 
    | >{\centering\arraybackslash}p{0.25\textwidth} 
    | >{\centering\arraybackslash}X | }
    
    \hline
    \textbf{Comparison} \textbf{Group} & \textbf{T-test results} \\
    \hline 
    RWL vs SRL & $t(49.1) = 0.022, p = 0.983,$ Cohen's $d = 0.006$ \\ 
    \hline
    RWL vs CRDAL & $t(40.6) = 0.213, p = 0.832,$ Cohen's $d = 0.055$ \\
    \hline
    SRL vs CRDAL & $t(52.7) = 0.266, p = 0.791,$ Cohen's $d = 0.069$ \\
    \hline
    
\end{tabularx}
\label{tab:t-test_step}
\end{table}

\subsection{Design-Space Exploration}

\begin{figure}[h]
    \centering
    \includegraphics[width=0.95\linewidth]{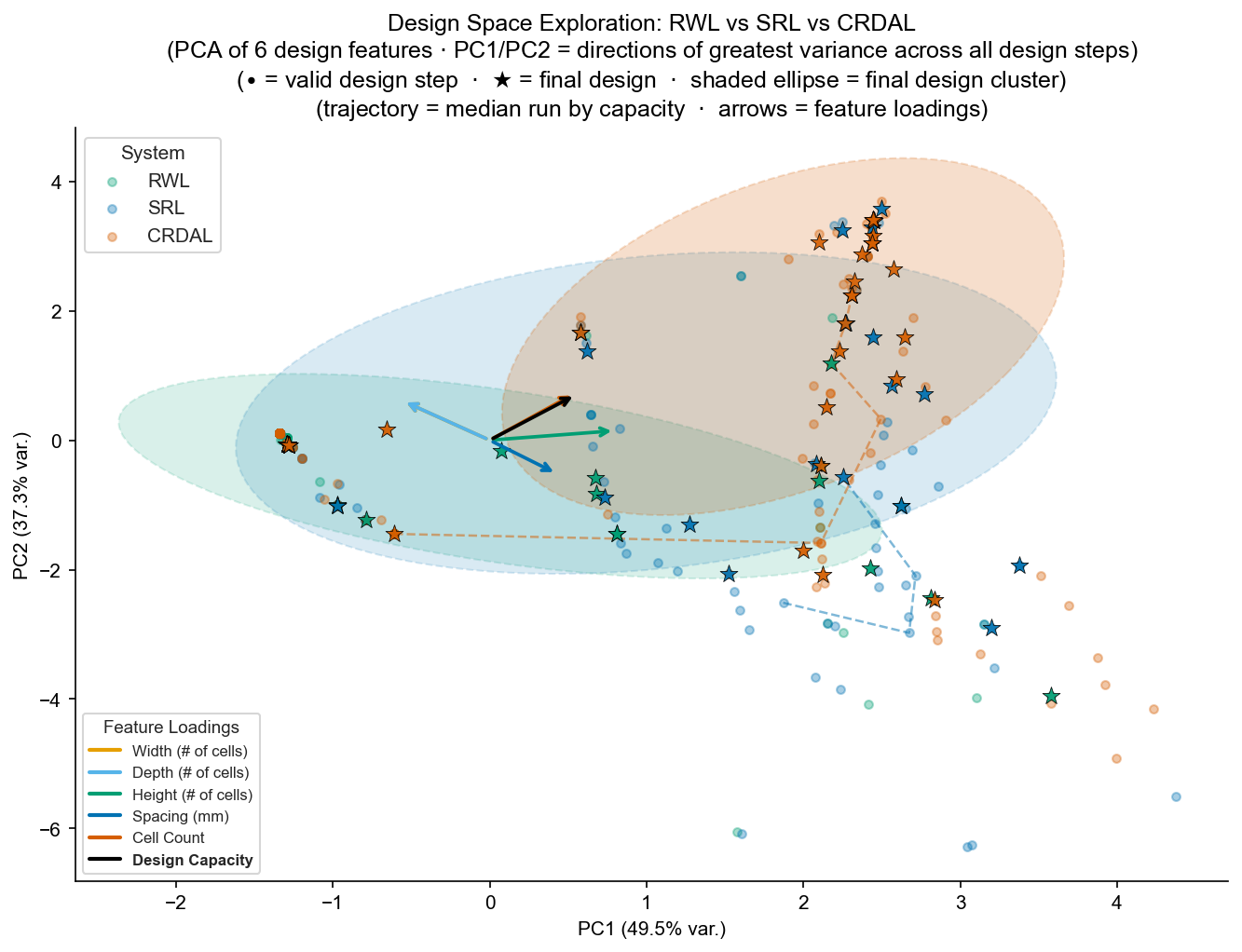}
    \caption{Design step trajectory and final design in the latent design space, explored by each agentic design system. }
    \label{fig:latent_space_pca}
\end{figure}

To better understand the differences in design-space exploration for the agentic systems, we examined the design-step trajectory and the final design in the latent design space, as illustrated in Figure~\ref{fig:latent_space_pca}. 
The three examined systems, RWL, SRL, and CRDAL, each explored and created designs that are different in the latent space. The final designs (the stars) from each system also occupy a different latent space (the ellipses). 

\begin{figure}[H]
    \centering
    \includegraphics[width=0.5\linewidth]{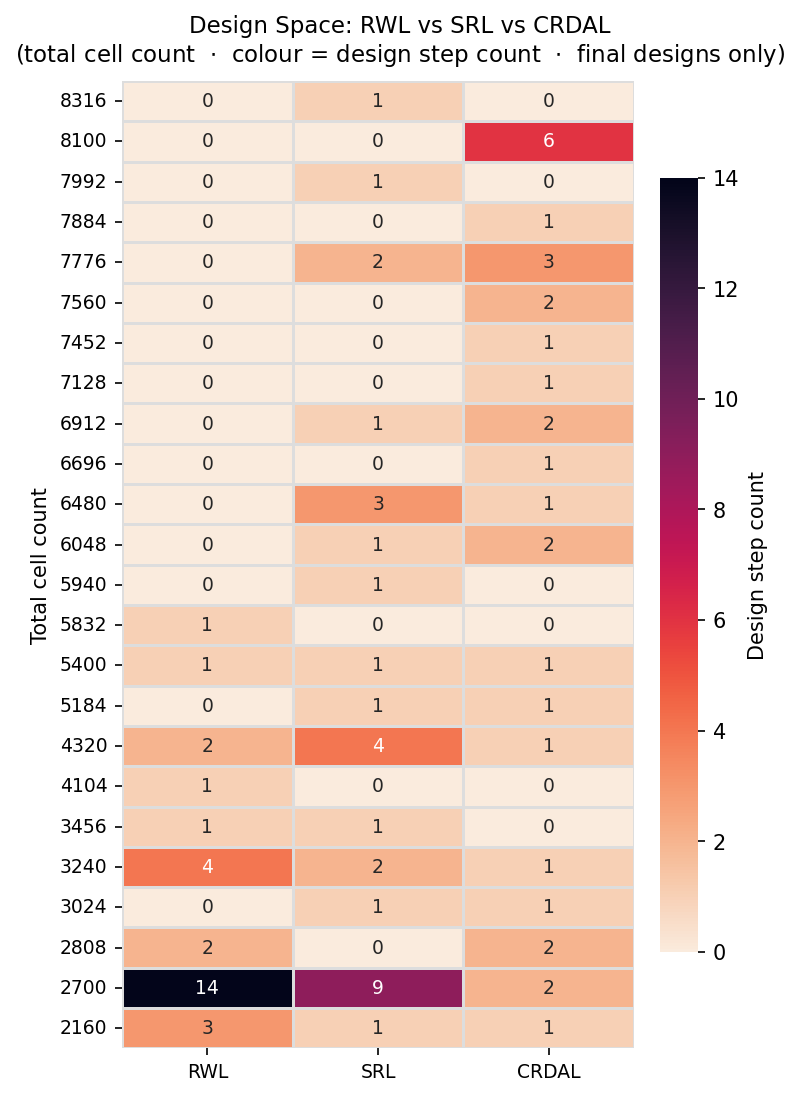}
    \caption{Cell series and parallel connections explored by each agentic design system. }
    \label{fig:cell_count_heatmap}
\end{figure}

Figure~\ref{fig:cell_count_heatmap} shows how the final designs vary in cell count between the systems. From the figure, the distributions of cell counts for the final designs across all three systems differ. All of the final designs from RWL have fewer than 6048 cells (29 out of 29). SRL shifted upward, with 9 of the 30 final designs having at least 6048 cells. However, most of the final designs from SRL have fewer than 6048 cells (21 out of 30). CRDAL further shifted substantially upward, with more than half (20 out of 30) final designs having more than 6048 cells. The cell count mode for the final designs from CRDAL is 8100 cells, with 6 out of 30 designs. 

\section{Discussion}

We discuss the results in terms of three themes: design performance, computational cost, and exploration behavior.

\subsection{Co-Regulation Design Agentic Loop (CRDAL) Creates Better Designs}

The proposed novel SRL and CRDAL created better designs than an RWL in the design problem tested in this paper. Further, the CRDAL system created better designs compared to SRL. The mean capacity of the battery pack design created by the CRDAL is significantly higher than that of the other two systems, with an average capacity of 141.17 Ah, compared to RWL's 106.08 Ah and SRL's 72.07 Ah (see Figure~\ref{fig:box_plot_RWL_SRL_CRDAL}).  This confirmed our second hypothesis that the novel metacognitive Co-Regulation Design Agentic Loop (CRDAL) improves system performance over the plain Ralph Wiggum Loop (RWL), and can create more optimal designs towards the given design objective ($t(39.4) = 7.174, p < 0.001,$ Cohen's $d = 1.846$). The CRDAL also performed significantly better than the SRL system ($t(57.8) = 2.888, p = 0.005,$ Cohen's $d = 0.746$). 

SRL, however, created the design with the highest capacity across all attempts from all three systems at $192.50 Ah$, matching the best design constructed by the authors. SRL also created better designs compared to RWL ($t(40.6) = 3.715, p = 0.001,$ Cohen's $d = 0.956$). This confirmed our first hypothesis that the novel metacognitive SRL improves system performance over the plain RWL, and can create more optimal designs towards the given design objective. 

Moreover, the computational cost of SRL and CRDAL is not statistically different from that for RWL (see Figure~\ref{fig:box_plot_RWL_SRL_CRDAL_steps}). Statistically speaking, the SRL and CRDAL created better designs without taking more design steps. This suggests that the SRL and CRDAL produce better-performing designs, not because they worked "harder" with more design attempts, but because they worked "smarter" through finding strategic methods to optimize. 

It is worth noting that all three systems examined in this work can effectively complete the given design task. All three systems generated valid final designs that satisfied design requirements and constraints, and optimized for the design objective in nearly all 30 runs, within the maximum allowed 30 design steps per run. Even for the base RWL, the lowest design capacity it created still satisfies the required minimum of $50Ah$. This shows that frontier LLMs can understand a multi-disciplinary engineering design problem, and attempt multi-step design optimization using specific design actions, reinforcing similar findings from other work in engineering design \cite{massoudi_agentic_2026, guo_toward_2025}. This also highlights the effectiveness of a Ralph Wiggum Loop, where an AI agent runs continuously in a loop and repeatedly attempts tasks until it succeeds, in engineering design tasks. The Ralph Wiggum Loop could be a simple and effective paradigm for agentic system architecture in other engineering design tasks. 

It is also interesting to see that CRDAL performed significantly better than SRL in our task. Both systems make use of the Progress Analyzer tool in each step that analyzes design step history and explicitly shows the design progress trajectory and trend summary. However, when there is another LLM agent that assesses the design progress and makes strategic suggestions for the next design iteration (as in CRDAL), the agentic system performs better than when the Design Agent is self-assessing and self-suggesting (as in SRL). Marvin Minsky suggested that intelligence comes from many minds, even "simple" ones, rather than a single "smart" mind \cite{minsky_society_1986}.  More modern research in engineering design has shown the benefits of using Multi-Agent Systems (MAS), where specialized agents collaborate to solve complex problems \cite{campbell_-design_1999, campbell_agent-based_2000, mccomb_drawing_2016, soria_zurita_design_2017}. Our finding resonates with this body of work. Specifically, our study suggests that even advanced modern reasoning LLM agents can still benefit from interacting and cooperating with other agents, and that the addition of a metacognitive assistant can greatly help the performance of the intelligent system. 

We anticipate that the results observed here relate to LLM context window dynamics. Prior studies suggest that LLMs perform the best when relevant information is stored at the very beginning or end of the context, regardless of context length \cite{liu_lost_2024, zhang_recursive_2026}. Perhaps the Metacognitive Co-Regulation Agent in CRDAL served as a fresh reminder to the Design Agent at each design step, keeping the Design Agent's attention sharp. More detailed analysis of the thinking traces is grounds for future work.


\subsection{Three Systems Explored Different Design Space}

As shown in Figure~\ref{fig:latent_space_pca}, the three examined systems explored and created designs that are different in the latent space. To unveil the difference in design reasoning that exists between these systems, we demonstrate one of the design trade-offs in the design problem used here. 

In the battery pack cell configuration design problem, one of the constraints is thermal. The battery pack should be capable of continuously supplying at least $48 A$ of current draw while staying at or below $60 ^{\circ} C$ during operation. In our simulation and design evaluation, the thermal dissipation rate is correlated with the surface area of the battery pack (assuming dominant convection and radiation with negligible conduction), and the temperature of the cells is related to both the heat generated by the cells and the dissipation rate. 

In the problem as framed, there are two ways to mitigate potential overheating issues: increasing the spacing between cells or adding more cells. Adding more cell spacing creates larger gaps between adjacent cells and allows more surface area for the pack. Therefore, it will lower the temperature. Adding more cells, although counterintuitive, also decreases the heat generated by the cells. We model the heat generated by each cell as 
\begin{equation}
    Q = I^2 \times R,
\end{equation}
where $Q$ is the heat generated by each cell, $I$ is the current carried by each cell, and $R$ is the internal resistance. Assuming a constant internal resistance, the heat generated by each cell is proportional to the current it carries, to the power of 2. That means lowering the current carried by each cell can very effectively reduce heat generation. Therefore, an effective way to lower the temperature of the cells is to use more cells and connect them in parallel (so the current $I$ is smaller) in the battery pack. Conveniently, having more cells connected in parallel also increases the capacity of the battery pack. 

However, adding more cells and increasing cell spacing both increase the footprint of the battery pack, which is constrained by physical dimensions, creating a design trade-off. Adding cell spacing would leave more empty space within the pack and reduce the number of cells it can fit, thereby also lowering capacity. Therefore, adding more cells in parallel connections is a much more effective way to mitigate the thermal issue than increasing cell spacing, as it greatly reduces heat generation while optimizing the design towards the objective (e.g., increasing capacity). 

We can see this effect in Figure~\ref{fig:latent_space_pca} as well. To gain the maximum possible design capacity, the designer should optimize towards the upper right corner of the figure along the direction of the arrow for capacity. The feature loading arrow for total cell count is almost parallel to the arrow for capacity. This suggests that adding more cells will effectively increase the design capacity. Also, the arrow for the number of cells in height is at a sharp angle to the arrow for capacity, also suggesting good effectiveness of adding more vertical layers. In contrast, the feature loading arrows for adding cell spacing and adding the number of cells in width and depth are at larger angles to the arrow for capacity, indicating less correlation. 

Based on Figure~\ref{fig:latent_space_pca}, the feature loading arrows suggest that CRDAL prioritized adding more cells (the orange ellipse), primarily by increasing the number of layers (design height), while RWL had more attempts at increasing the cell spacing (the green ellipse). CRDAL explores the upper right corner more effectively in the latent design space. The SRL (the blue ellipse) is somewhere in between RWL and CRDAL. Given the similar locations in latent design space for the middle steps, it might also suggest that CRDAL can escape local minima more effectively. Overall, this partially explains why the CRDAL system generated designs with higher capacity. This also partially supports our third hypothesis, that the metacognitive Co-Regulation Design Agent Loop mitigates design fixation, and allows the agentic system to explore design space more effectively. 

The distributions of cell counts for the final designs from the three systems shown in Figure~\ref{fig:cell_count_heatmap} show how the designs differ in cell count across systems. CRDAL is the only system that consistently explored battery packs with more than 6048 cells. Assuming that's with $400 V / 3.7 V \approx 108$ cells in parallel per series, that leads to at least $6048 / 108 = 56$ series connections and at least $56 \times 2.5 Ah = 140 Ah$ capacity. In fact, the cell count mode for the final designs from CRDAL is 8100 cells ($\approx 187.5 Ah$), with 6 out of 30 designs. In comparison, RWL and SRL each have 0 and 9 final designs, respectively, with more than 6048 cells across the 29 and 30 successful runs. This suggests that CRDAL more effectively explored the design space with more cells, ultimately leading to higher capacity.

\subsection{Limitations and Future Work}

This work faces several limitations. First, we examined the agent design systems in a design problem on battery pack cell configuration. It is a multi-disciplinary engineering design problem with multi-step design optimization and objective design performance evaluation, which we believe to be suitable for demonstrating the performance difference between the agentic systems in this paper. However, future work should explore engineering design problems in different engineering domains and with various complexities, design objectives, and constraints, including non-convex Pareto fronts and qualitative design decisions.  

Second, in this work, we used a frontier LLM, Gemini 3.1 Pro. It was chosen for its best-in-class performance in math and reasoning at the time of writing. However, different LLMs may exhibit different behavior and performance, potentially influencing agentic system performance. Future work should further examine agentic systems with other LLMs, potentially including fine-tuned LLMs for engineering design or design reasoning. It is also worth investigating with smaller models, such as Qwen 3.5, for better local deployability. 

Third, in this work, we limited the tool call function to only the Numerical Evaluator and Validator (and Progress Analyzer for SRL and CRDAL), without providing the Design Agent with tools that could be used more directly to optimize designs. This is to accentuate the Design Agent's design reasoning and strategy-making, as well as the interaction between agents. Future work should further test design agents with different advanced tool-call functions (e.g., numerical optimizers, web search, spatial design representation generation) for more capable agentic AI systems for real-world applications. 

In addition, future work should explore multi-agent agentic AI systems with multiple design agents and/or multiple metacognitive support agents, as research has suggested that Multi-Agent Systems could be more effective for solving complex, parallel, and multi-disciplinary problems \cite{wooldridge_intelligent_1995, campbell_-design_1999, campbell_agent-based_2000, stone_multiagent_2000, gerhard_distributed_2001, mccomb_drawing_2016, soria_zurita_design_2017, rismiller_adversarial_2021, rismiller_exploring_2023, massoudi_agentic_2026, chen_llm-based_2026}. A deeper understanding of interactions and metacognition among multiple agents can facilitate the development of complex agentic AI systems.

\section{Conclusion}

In this work, we propose (1) a novel Self-Regulation Loop (SRL), in which the Design Agent self-regulates and explicitly monitors its own metacognition, and (2) a novel Co-Regulation Design Agentic Loop (CRDAL), in which a Metacognitive Co-Regulation Agent assists a Design Agent in metacognition to mitigate design fixation, thereby improving system performance for engineering design tasks. In the battery pack design problem examined, we found that the novel SRL and CRDAL systems generate designs with better performance, without significantly increasing the computational cost, compared to a plain Ralph Wiggum Loop (RWL). Further, the novel CRDAL generates designs with significantly better performance than SRL. Also, we found that the CRDAL system navigated through the latent design space more effectively. The proposed system architectures and findings of this work provide practical implications for future development of agentic AI systems for engineering design.

\begin{ack}

This material is based upon work supported by the National Science Foundation under Grant No. 2118924 Supporting Designers in Learning to Co-create with AI for Complex Computational Design Tasks. This work was also supported by the Collaborative Grand Challenge Initiative by the Department of Mechanical Engineering at Carnegie Mellon University. This research is also conducted with support from Google.org and the Google Cloud Research Credits program for the Gemini Academic Program.

\end{ack}

\bibliographystyle{plainnat}
\bibliography{references}

@misc{guo_toward_2025,
	title = {Toward {Engineering} {AGI}: {Benchmarking} the {Engineering} {Design} {Capabilities} of {LLMs}},
	shorttitle = {Toward {Engineering} {AGI}},
	url = {http://arxiv.org/abs/2509.16204},
	doi = {10.48550/arXiv.2509.16204},
	urldate = {2026-03-23},
	publisher = {arXiv},
	author = {Guo, Xingang and Li, Yaxin and Kong, Xiangyi and Jiang, Yilan and Zhao, Xiayu and Gong, Zhihua and Zhang, Yufan and Li, Daixuan and Sang, Tianle and Zhu, Beixiao and Jun, Gregory and Huang, Yingbing and Liu, Yiqi and Xue, Yuqi and Kundu, Rahul Dev and Lim, Qi Jian and Zhao, Yizhou and Granger, Luke Alexander and Younis, Mohamed Badr and Keivan, Darioush and Sabharwal, Nippun and Sinha, Shreyanka and Agarwal, Prakhar and Vandyck, Kojo and Mai, Hanlin and Wang, Zichen and Venkatesh, Aditya and Barik, Ayush and Yang, Jiankun and Yue, Chongying and He, Jingjie and Wang, Libin and Xu, Licheng and Chen, Hao and Wang, Jinwen and Xu, Liujun and Shetty, Rushabh and Guo, Ziheng and Song, Dahui and Jha, Manvi and Liang, Weijie and Yan, Weiman and Zhang, Bryan and Karnoor, Sahil Bhandary and Zhang, Jialiang and Pandya, Rutva and Gong, Xinyi and Ganesh, Mithesh Ballae and Shi, Feize and Xu, Ruiling and Zhang, Yifan and Ouyang, Yanfeng and Qin, Lianhui and Rosenbaum, Elyse and Snyder, Corey and Seiler, Peter and Dullerud, Geir and Zhang, Xiaojia Shelly and Cheng, Zuofu and Hanumolu, Pavan Kumar and Huang, Jian and Kulkarni, Mayank and Namazifar, Mahdi and Zhang, Huan and Hu, Bin},
	month = nov,
	year = {2025},
	note = {arXiv:2509.16204 [cs]},
}

@misc{noauthor_gemini_2026,
	title = {Gemini 3.1 {Pro}: {A} smarter model for your most complex tasks},
	shorttitle = {Gemini 3.1 {Pro}},
	url = {https://blog.google/innovation-and-ai/models-and-research/gemini-models/gemini-3-1-pro/},
	language = {en-us},
	urldate = {2026-03-23},
	journal = {Google},
	month = feb,
	year = {2026},
}

@article{grecu_design_1996,
	title = {Design agents that learn},
	volume = {10},
	issn = {1469-1760, 0890-0604},
	url = {https://www.cambridge.org/core/journals/ai-edam/article/abs/design-agents-that-learn/C604E8564A369D4E664D6BF4F02157A7},
	doi = {10.1017/S0890060400001426},
	language = {en},
	number = {2},
	urldate = {2026-03-23},
	journal = {AI EDAM},
	author = {Grecu, Dan L. and Brown, David C.},
	month = apr,
	year = {1996},
	pages = {149--150},
}

@article{mccomb_optimizing_2017,
	title = {Optimizing {Design} {Teams} {Based} on {Problem} {Properties}: {Computational} {Team} {Simulations} and an {Applied} {Empirical} {Test}},
	volume = {139},
	issn = {1050-0472},
	shorttitle = {Optimizing {Design} {Teams} {Based} on {Problem} {Properties}},
	url = {https://doi.org/10.1115/1.4035793},
	doi = {10.1115/1.4035793},
	number = {041101},
	urldate = {2026-03-22},
	journal = {Journal of Mechanical Design},
	author = {McComb, Christopher and Cagan, Jonathan and Kotovsky, Kenneth},
	month = feb,
	year = {2017},
}

@article{moss_learning_2004,
	title = {Learning from design experience in an agent-based design system},
	volume = {15},
	issn = {1435-6066},
	url = {https://doi.org/10.1007/s00163-003-0042-4},
	doi = {10.1007/s00163-003-0042-4},
	language = {en},
	number = {2},
	urldate = {2026-03-22},
	journal = {Research in Engineering Design},
	author = {Moss, Jarrod and Cagan, Jonathan and Kotovsky, Kenneth},
	month = sep,
	year = {2004},
	pages = {77--92},
}

@article{yuan_experimental_2018,
	title = {Experimental {Study} on the {Associations} {Among} {Sketches} {Based} on {Design} {Cognition}},
	volume = {140},
	issn = {1050-0472},
	url = {https://doi.org/10.1115/1.4040627},
	doi = {10.1115/1.4040627},
	number = {101102},
	urldate = {2026-03-22},
	journal = {Journal of Mechanical Design},
	author = {Yuan, Ping and Li, Yan and Chen, Jian and Xiong, Yan and Liu, Longfan},
	month = jul,
	year = {2018},
}

@article{brown_study_2024,
	title = {A {Study} on {Generative} {Design} {Reasoning} and {Students}' {Divergent} and {Convergent} {Thinking}},
	volume = {146},
	issn = {1050-0472},
	url = {https://doi.org/10.1115/1.4064564},
	doi = {10.1115/1.4064564},
	number = {031405},
	urldate = {2026-03-22},
	journal = {Journal of Mechanical Design},
	author = {Brown, Alex and Goldstein, Molly H. and Clay, John and Demirel, H. Onan and Li, Xingang and Sha, Zhenghui},
	month = feb,
	year = {2024},
}

@article{raina_learning_2019,
	title = {Learning to {Design} {From} {Humans}: {Imitating} {Human} {Designers} {Through} {Deep} {Learning}},
	volume = {141},
	issn = {1050-0472},
	shorttitle = {Learning to {Design} {From} {Humans}},
	url = {https://doi.org/10.1115/1.4044256},
	doi = {10.1115/1.4044256},
	number = {111102},
	urldate = {2026-03-22},
	journal = {Journal of Mechanical Design},
	author = {Raina, Ayush and McComb, Christopher and Cagan, Jonathan},
	month = sep,
	year = {2019},
}

@article{stahovich_learnit_1999,
	title = {{LearnIT}: {An} {Instance}-{Based} {Approach} to {Learning} and {Reusing} {Design} {Strategies}},
	volume = {122},
	issn = {1050-0472},
	shorttitle = {{LearnIT}},
	url = {https://doi.org/10.1115/1.1288216},
	doi = {10.1115/1.1288216},
	number = {3},
	urldate = {2026-03-22},
	journal = {Journal of Mechanical Design},
	author = {Stahovich, Thomas F.},
	month = sep,
	year = {1999},
	pages = {249--256},
}

@article{rahman_predicting_2021,
	title = {Predicting {Sequential} {Design} {Decisions} {Using} the {Function}-{Behavior}-{Structure} {Design} {Process} {Model} and {Recurrent} {Neural} {Networks}},
	volume = {143},
	issn = {1050-0472},
	url = {https://doi.org/10.1115/1.4049971},
	doi = {10.1115/1.4049971},
	number = {081706},
	urldate = {2026-03-22},
	journal = {Journal of Mechanical Design},
	author = {Rahman, Molla Hafizur and Xie, Charles and Sha, Zhenghui},
	month = mar,
	year = {2021},
}

@article{allison_special_2022,
	title = {Special {Issue}: {Artificial} {Intelligence} and {Engineering} {Design}},
	volume = {144},
	issn = {1050-0472},
	shorttitle = {Special {Issue}},
	url = {https://doi.org/10.1115/1.4053111},
	doi = {10.1115/1.4053111},
	number = {020301},
	urldate = {2026-03-22},
	journal = {Journal of Mechanical Design},
	editor = {Allison, James T. and Cardin, Michel-Alexandre and McComb, Chris and Ren, Max Yi and Selva, Daniel and Tucker, Conrad and Witherell, Paul and Zhao, Yaoyao Fiona},
	month = jan,
	year = {2022},
}

@article{rismiller_exploring_2023,
	title = {Exploring the impact of set-based concurrent engineering through multi-agent system simulation},
	volume = {37},
	issn = {0890-0604, 1469-1760},
	doi = {10.1017/S0890060423000112},
	language = {en},
	urldate = {2026-03-22},
	journal = {AI EDAM},
	author = {Rismiller, Sean and Cagan, Jonathan and McComb, Christopher},
	month = jan,
	year = {2023},
	pages = {e16},
}

@article{campbell_agent-based_2000,
	title = {Agent-{Based} {Synthesis} of {Electromechanical} {Design} {Configurations}},
	volume = {122},
	issn = {1050-0472},
	url = {https://doi.org/10.1115/1.533546},
	doi = {10.1115/1.533546},
	number = {1},
	urldate = {2026-03-22},
	journal = {Journal of Mechanical Design},
	author = {Campbell, Matthew I. and Cagan, Jonathan and Kotovsky, Kenneth},
	month = jan,
	year = {2000},
	pages = {61--69},
}

@article{chen_llm-based_2026,
	title = {An {LLM}-based multi-agent system to assist early-stage product design and evaluation},
	volume = {37},
	issn = {0954-4828},
	url = {https://doi.org/10.1080/09544828.2026.2616583},
	doi = {10.1080/09544828.2026.2616583},
	number = {3},
	urldate = {2026-03-22},
	journal = {Journal of Engineering Design},
	publisher = {Taylor \& Francis},
	author = {Chen, Pei and Cai, Yichen and Zhou, Zihong and Yao, Jiayi and Li, Jiayang and You, Weitao and Sun, Lingyun},
	month = mar,
	year = {2026},
	note = {\_eprint: https://doi.org/10.1080/09544828.2026.2616583},
	pages = {945--980},
}

@article{rismiller_adversarial_2021,
	title = {An {Adversarial} {Agent}-{Based} {Design} {Method} {Using} {Stochastic} {Stackelberg} {Game} {Conditions}},
	volume = {143},
	issn = {1050-0472},
	url = {https://doi.org/10.1115/1.4049862},
	doi = {10.1115/1.4049862},
	number = {031714},
	urldate = {2026-03-22},
	journal = {Journal of Mechanical Design},
	author = {Rismiller, Sean C. and Cagan, Jonathan and McComb, Christopher},
	month = jan,
	year = {2021},
}

@article{mccomb_drawing_2016,
	title = {Drawing {Inspiration} {From} {Human} {Design} {Teams} for {Better} {Search} and {Optimization}: {The} {Heterogeneous} {Simulated} {Annealing} {Teams} {Algorithm}},
	volume = {138},
	issn = {1050-0472},
	shorttitle = {Drawing {Inspiration} {From} {Human} {Design} {Teams} for {Better} {Search} and {Optimization}},
	url = {https://doi.org/10.1115/1.4032810},
	doi = {10.1115/1.4032810},
	number = {044501},
	urldate = {2026-03-22},
	journal = {Journal of Mechanical Design},
	author = {McComb, Christopher and Cagan, Jonathan and Kotovsky, Kenneth},
	month = mar,
	year = {2016},
}

@article{soria_zurita_design_2017,
	title = {Design of {Complex} {Engineered} {Systems} {Using} {Multi}-{Agent} {Coordination}},
	volume = {18},
	issn = {1530-9827},
	url = {https://doi.org/10.1115/1.4038158},
	doi = {10.1115/1.4038158},
	number = {011003},
	urldate = {2026-03-22},
	journal = {Journal of Computing and Information Science in Engineering},
	author = {Soria Zurita, Nicolás F. and Colby, Mitchell K. and Tumer, Irem Y. and Hoyle, Christopher and Tumer, Kagan},
	month = nov,
	year = {2017},
}

@article{gerhard_distributed_2001,
	title = {A {Distributed} {Product} {Realization} {Environment} for {Design} and {Manufacturing}},
	volume = {1},
	issn = {1530-9827},
	url = {https://doi.org/10.1115/1.1412230},
	doi = {10.1115/1.1412230},
	number = {3},
	urldate = {2026-03-22},
	journal = {Journal of Computing and Information Science in Engineering},
	author = {Gerhard, Jonathan F. and Rosen, David and Allen, Janet K. and Mistree, Farrokh},
	month = aug,
	year = {2001},
	pages = {235--244},
}

@article{campbell_-design_1999,
	title = {A-{Design}: {An} {Agent}-{Based} {Approach} to {Conceptual} {Design} in a {Dynamic} {Environment}},
	volume = {11},
	issn = {1435-6066},
	shorttitle = {A-{Design}},
	url = {https://doi.org/10.1007/s001630050013},
	doi = {10.1007/s001630050013},
	language = {en},
	number = {3},
	urldate = {2026-03-22},
	journal = {Research in Engineering Design},
	author = {Campbell, Matthew I. and Cagan, Jonathan and Kotovsky, Kenneth},
	month = oct,
	year = {1999},
	pages = {172--192},
}

@article{stone_multiagent_2000,
	title = {Multiagent {Systems}: {A} {Survey} from a {Machine} {Learning} {Perspective}},
	volume = {8},
	issn = {1573-7527},
	url = {https://doi.org/10.1023/A:1008942012299},
	doi = {10.1023/A:1008942012299},
	number = {3},
	journal = {Autonomous Robots},
	author = {Stone, Peter and Veloso, Manuela},
	month = jun,
	year = {2000},
	pages = {345--383},
}

@article{wooldridge_intelligent_1995,
	title = {Intelligent agents: theory and practice},
	volume = {10},
	url = {https://api.semanticscholar.org/CorpusID:221342993},
	journal = {The Knowledge Engineering Review},
	author = {Wooldridge, Michael and Jennings, Nicholas R.},
	year = {1995},
	pages = {115 -- 152},
}

@inproceedings{zhou_metagent-p_2025,
	address = {Vienna, Austria},
	title = {Metagent-{P}: {A} {Neuro}-{Symbolic} {Planning} {Agent} with {Metacognition} for {Open} {Worlds}},
	isbn = {979-8-89176-256-5},
	url = {https://aclanthology.org/2025.findings-acl.1169/},
	doi = {10.18653/v1/2025.findings-acl.1169},
	booktitle = {Findings of the {Association} for {Computational} {Linguistics}: {ACL} 2025},
	publisher = {Association for Computational Linguistics},
	author = {Zhou, Yanfang and Liu, Yuntao and Li, Xiaodong and Zhao, Yongqiang and Wang, Xintong and Tian, Jinlong and Li, Zhenyu and Xu, Xinhai},
	editor = {Che, Wanxiang and Nabende, Joyce and Shutova, Ekaterina and Pilehvar, Mohammad Taher},
	month = jul,
	year = {2025},
	pages = {22747--22764},
}

@inproceedings{renze_self-reflection_2024,
	title = {Self-{Reflection} in {Large} {Language} {Model} {Agents}: {Effects} on {Problem}-{Solving} {Performance}},
	doi = {10.1109/FLLM63129.2024.10852426},
	booktitle = {2024 2nd {International} {Conference} on {Foundation} and {Large} {Language} {Models} ({FLLM})},
	author = {Renze, Matthew and Guven, Erhan},
	year = {2024},
	pages = {516--525},
}

@inproceedings{wang_metacognitive_2024,
	address = {Mexico City, Mexico},
	title = {Metacognitive {Prompting} {Improves} {Understanding} in {Large} {Language} {Models}},
	url = {https://aclanthology.org/2024.naacl-long.106/},
	doi = {10.18653/v1/2024.naacl-long.106},
	booktitle = {Proceedings of the 2024 {Conference} of the {North} {American} {Chapter} of the {Association} for {Computational} {Linguistics}: {Human} {Language} {Technologies} ({Volume} 1: {Long} {Papers})},
	publisher = {Association for Computational Linguistics},
	author = {Wang, Yuqing and Zhao, Yun},
	editor = {Duh, Kevin and Gomez, Helena and Bethard, Steven},
	month = jun,
	year = {2024},
	pages = {1914--1926},
}

@inproceedings{liu_position_2025,
	address = {Vancouver, BC, Canada},
	title = {Position: {Truly} {Self}-{Improving} {Agents} {Require} {Intrinsic} {Metacognitive} {Learning}},
	copyright = {Creative Commons Attribution 4.0 International},
	doi = {10.48550/ARXIV.2506.05109},
	urldate = {2026-03-22},
	booktitle = {Proceedings of the 42nd {International} {Conference} on {Machine} {Learning}},
	publisher = {PMLR},
	author = {Liu, Tennison and van der Schaar, Mihaela},
	year = {2025},
	note = {Version Number: 1},
}

@incollection{brown_knowing_1978,
	address = {Hillsdale, NJ},
	title = {Knowing {When}, {Where}, and {How} to {Remember}: {A} {Problem} of {Metacognition}},
	volume = {1},
	booktitle = {Advances in {Instructional} {Psychology}},
	publisher = {Lawrence Erlbaum Associates},
	author = {Brown, Ann L.},
	editor = {Glaser, Robert},
	year = {1978},
	pages = {77--165},
}

@book{kellogg_cognitive_2002,
	edition = {Second Edition},
	title = {Cognitive {Psychology}},
	isbn = {978-0-7619-2130-1},
	url = {https://us.sagepub.com/en-us/nam/cognitive-psychology/book10816},
	language = {en},
	urldate = {2026-03-22},
	publisher = {SAGE Publications, Inc},
	author = {Kellogg, Ronald T.},
	month = aug,
	year = {2002},
}

@article{flavell_metacognition_1979,
	address = {US},
	title = {Metacognition and cognitive monitoring: {A} new area of cognitive–developmental inquiry},
	volume = {34},
	issn = {1935-990X},
	shorttitle = {Metacognition and cognitive monitoring},
	doi = {10.1037/0003-066X.34.10.906},
	number = {10},
	journal = {American Psychologist},
	publisher = {American Psychological Association},
	author = {Flavell, John H.},
	year = {1979},
	pages = {906--911},
}

@inproceedings{dixon_use_2012,
	title = {The {Use} of {Executive} {Control} {Processes} in {Engineering} {Design} by {Engineering} {Students} and {Professional} {Engineers}},
	volume = {24},
	issn = {1045-1064},
	url = {http://scholar.lib.vt.edu/ejournals/JTE/v24n1/dixon2.html},
	doi = {10.21061/jte.v24i1.a.5},
	number = {1},
	urldate = {2026-03-22},
	booktitle = {Journal of {Technology} {Education}},
	author = {Dixon, Raymond A. and Johnson, Scott D.},
	month = sep,
	year = {2012},
	note = {Journal Abbreviation: JTE},
}

@article{pegel_pareto-optimal_2025,
	title = {Pareto-{Optimal} {Design} of {Automotive} {Battery} {Systems} with {Tabless} {Cylindrical} {Lithium}-{Ion} {Cells}: {Resolving} the {Trade}-{Off} {Between} {Energy}, {Performance}, {Weight}, and {Cost} for {Variable} {Vehicle} {Requirements}},
	volume = {13},
	copyright = {© 2024 The Author(s). Energy Technology published by Wiley-VCH GmbH},
	issn = {2194-4296},
	shorttitle = {Pareto-{Optimal} {Design} of {Automotive} {Battery} {Systems} with {Tabless} {Cylindrical} {Lithium}-{Ion} {Cells}},
	url = {https://onlinelibrary.wiley.com/doi/abs/10.1002/ente.202401479},
	doi = {10.1002/ente.202401479},
	language = {en},
	number = {4},
	urldate = {2026-03-21},
	journal = {Energy Technology},
	author = {Pegel, Hendrik and Jany, Lukas and Sauer, Dirk Uwe},
	year = {2025},
	note = {\_eprint: https://onlinelibrary.wiley.com/doi/pdf/10.1002/ente.202401479},
	pages = {2401479},
}

@article{balasubramanian_study_2025,
	title = {Study on the battery thermal management system for cylindrical lithium-ion battery with nano-doped phase change material and liquid cooling},
	volume = {15},
	copyright = {2025 The Author(s)},
	issn = {2045-2322},
	url = {https://www.nature.com/articles/s41598-025-08884-5},
	doi = {10.1038/s41598-025-08884-5},
	language = {en},
	number = {1},
	urldate = {2026-03-21},
	journal = {Scientific Reports},
	publisher = {Nature Publishing Group},
	author = {Balasubramanian, Dhinesh and Venugopal, Inbanaathan Papla and Subramanian, Mohankumar and Raja, Vijayanandh and Kale, Utku and Matijošius, Jonas},
	month = jul,
	year = {2025},
	pages = {24053},
}

@misc{noauthor_gemini_nodate,
	title = {Gemini 3 {Developer} {Guide} {\textbar} {Gemini} {API}},
	url = {https://ai.google.dev/gemini-api/docs/gemini-3},
	language = {en},
	urldate = {2026-03-21},
	journal = {Google AI for Developers},
}

@article{valkenburg_reflective_1998,
	title = {The reflective practice of design teams},
	volume = {19},
	issn = {0142-694X},
	url = {https://www.sciencedirect.com/science/article/pii/S0142694X98000118},
	doi = {10.1016/S0142-694X(98)00011-8},
	number = {3},
	urldate = {2026-03-21},
	journal = {Design Studies},
	author = {Valkenburg, Rianne and Dorst, Kees},
	month = jul,
	year = {1998},
	pages = {249--271},
}

@inproceedings{roozenburg_describing_1998,
	address = {London},
	title = {Describing {Design} as a {Reflective} {Practice}: {Observations} on {Schön}’s {Theory} of {Practice}},
	isbn = {978-1-4471-1268-6},
	shorttitle = {Describing {Design} as a {Reflective} {Practice}},
	doi = {10.1007/978-1-4471-1268-6_3},
	language = {en},
	booktitle = {Designers},
	publisher = {Springer},
	author = {Roozenburg, Norbert F. M. and Dorst, Kees},
	editor = {Frankenberger, Eckart and Birkhofer, Herbert and Badke-Schaub, Petra},
	year = {1998},
	pages = {29--41},
}

@article{schon_designing_1992,
	series = {Artificial {Intelligence} in {Design} {Conference} 1991 {Special} {Issue}},
	title = {Designing as reflective conversation with the materials of a design situation},
	volume = {5},
	issn = {0950-7051},
	url = {https://www.sciencedirect.com/science/article/pii/095070519290020G},
	doi = {10.1016/0950-7051(92)90020-G},
	number = {1},
	urldate = {2026-03-21},
	journal = {Knowledge-Based Systems},
	author = {Schön, D. A.},
	month = mar,
	year = {1992},
	pages = {3--14},
}

@book{schon_reflective_2017,
	address = {London},
	title = {The {Reflective} {Practitioner}: {How} {Professionals} {Think} in {Action}},
	isbn = {978-1-315-23747-3},
	shorttitle = {The {Reflective} {Practitioner}},
	doi = {10.4324/9781315237473},
	publisher = {Routledge},
	author = {Schön, Donald A.},
	month = mar,
	year = {2017},
}

@misc{huntley_everything_2026,
	title = {everything is a ralph loop},
	url = {https://ghuntley.com/loop/},
	language = {en},
	urldate = {2026-03-21},
	journal = {Geoffrey Huntley},
	author = {Huntley, Geoffrey},
	month = jan,
	year = {2026},
}

@inproceedings{wei_chain--thought_2022,
	address = {Red Hook, NY, USA},
	series = {{NIPS} '22},
	title = {Chain-of-thought prompting elicits reasoning in large language models},
	isbn = {978-1-7138-7108-8},
	booktitle = {Proceedings of the 36th {International} {Conference} on {Neural} {Information} {Processing} {Systems}},
	publisher = {Curran Associates Inc.},
	author = {Wei, Jason and Wang, Xuezhi and Schuurmans, Dale and Bosma, Maarten and Ichter, Brian and Xia, Fei and Chi, Ed H. and Le, Quoc V. and Zhou, Denny},
	year = {2022},
}

@inproceedings{brown_language_2020,
	address = {Red Hook, NY, USA},
	series = {{NIPS} '20},
	title = {Language models are few-shot learners},
	isbn = {978-1-7138-2954-6},
	booktitle = {Proceedings of the 34th {International} {Conference} on {Neural} {Information} {Processing} {Systems}},
	publisher = {Curran Associates Inc.},
	author = {Brown, Tom B. and Mann, Benjamin and Ryder, Nick and Subbiah, Melanie and Kaplan, Jared and Dhariwal, Prafulla and Neelakantan, Arvind and Shyam, Pranav and Sastry, Girish and Askell, Amanda and Agarwal, Sandhini and Herbert-Voss, Ariel and Krueger, Gretchen and Henighan, Tom and Child, Rewon and Ramesh, Aditya and Ziegler, Daniel M. and Wu, Jeffrey and Winter, Clemens and Hesse, Christopher and Chen, Mark and Sigler, Eric and Litwin, Mateusz and Gray, Scott and Chess, Benjamin and Clark, Jack and Berner, Christopher and McCandlish, Sam and Radford, Alec and Sutskever, Ilya and Amodei, Dario},
	year = {2020},
}

@inproceedings{kojima_large_2022,
	address = {Red Hook, NY, USA},
	series = {{NIPS} '22},
	title = {Large language models are zero-shot reasoners},
	isbn = {978-1-7138-7108-8},
	booktitle = {Proceedings of the 36th {International} {Conference} on {Neural} {Information} {Processing} {Systems}},
	publisher = {Curran Associates Inc.},
	author = {Kojima, Takeshi and Gu, Shixiang Shane and Reid, Machel and Matsuo, Yutaka and Iwasawa, Yusuke},
	year = {2022},
}

@article{zoph_emergent_2022,
	title = {Emergent abilities of large language models},
	journal = {TMLR},
	author = {Zoph, Barret and Raffel, Colin and Schuurmans, Dale and Yogatama, Dani and Zhou, Denny and Metzler, Don and Chi, Ed H. and Wei, Jason and Dean, Jeff and Fedus, Liam B. and Bosma, Maarten Paul and Vinyals, Oriol and Liang, Percy and Borgeaud, Sebastian and Hashimoto, Tatsunori B. and Tay, Yi},
	year = {2022},
}

@article{linsey_study_2010,
	title = {A {Study} of {Design} {Fixation}, {Its} {Mitigation} and {Perception} in {Engineering} {Design} {Faculty}},
	volume = {132},
	issn = {1050-0472},
	url = {https://doi.org/10.1115/1.4001110},
	doi = {10.1115/1.4001110},
	number = {041003},
	urldate = {2026-03-20},
	journal = {Journal of Mechanical Design},
	author = {Linsey, J. S. and Tseng, I. and Fu, K. and Cagan, J. and Wood, K. L. and Schunn, C.},
	month = apr,
	year = {2010},
}

@article{viswanathan_design_2013,
	title = {Design {Fixation} and {Its} {Mitigation}: {A} {Study} on the {Role} of {Expertise}},
	volume = {135},
	issn = {1050-0472},
	shorttitle = {Design {Fixation} and {Its} {Mitigation}},
	url = {https://doi.org/10.1115/1.4024123},
	doi = {10.1115/1.4024123},
	number = {051008},
	urldate = {2026-03-20},
	journal = {Journal of Mechanical Design},
	author = {Viswanathan, Vimal K. and Linsey, Julie S.},
	month = apr,
	year = {2013},
}

@article{jansson_design_1991,
	title = {Design fixation},
	volume = {12},
	issn = {0142-694X},
	url = {https://www.sciencedirect.com/science/article/pii/0142694X9190003F},
	doi = {10.1016/0142-694X(91)90003-F},
	number = {1},
	urldate = {2026-03-20},
	journal = {Design Studies},
	author = {Jansson, David G. and Smith, Steven M.},
	month = jan,
	year = {1991},
	pages = {3--11},
}

@misc{chen_understanding_2025,
	title = {Understanding {Design} {Fixation} in {Generative} {AI}},
	url = {http://arxiv.org/abs/2502.05870},
	doi = {10.48550/arXiv.2502.05870},
	urldate = {2026-03-20},
	publisher = {arXiv},
	author = {Chen, Liuqing and Song, Yaxuan and Zheng, Chunyuan and Jing, Qianzhi and Hansen, Preben and Sun, Lingyun},
	month = feb,
	year = {2025},
	note = {arXiv:2502.05870 [cs]},
}

@incollection{mccaslin_self-regulated_2001,
	address = {Mahwah, NJ, US},
	title = {Self-regulated learning and academic achievement: {A} {Vygotskian} view},
	isbn = {978-0-8058-3560-1 978-0-8058-3561-8},
	shorttitle = {Self-regulated learning and academic achievement},
	booktitle = {Self-regulated learning and academic achievement: {Theoretical} perspectives, 2nd ed},
	publisher = {Lawrence Erlbaum Associates Publishers},
	author = {McCaslin, Mary and Hickey, Daniel T.},
	year = {2001},
	pages = {227--252},
}

@article{allal_assessment_2020,
	title = {Assessment and the co-regulation of learning in the classroom},
	volume = {27},
	issn = {0969-594X},
	url = {https://doi.org/10.1080/0969594X.2019.1609411},
	doi = {10.1080/0969594X.2019.1609411},
	number = {4},
	urldate = {2026-03-20},
	journal = {Assessment in Education: Principles, Policy \& Practice},
	publisher = {Routledge},
	author = {Allal, Linda},
	month = jul,
	year = {2020},
	note = {\_eprint: https://doi.org/10.1080/0969594X.2019.1609411},
	pages = {332--349},
}

@article{lim_co-regulation_2020,
	title = {Co-regulation in collaborative learning: {Grounded} in achievement goal theory},
	volume = {103},
	issn = {0883-0355},
	shorttitle = {Co-regulation in collaborative learning},
	url = {https://www.sciencedirect.com/science/article/pii/S0883035519312327},
	doi = {10.1016/j.ijer.2020.101621},
	urldate = {2026-03-20},
	journal = {International Journal of Educational Research},
	author = {Lim, Ji Young and Lim, Kyu Yon},
	month = jan,
	year = {2020},
	pages = {101621},
}

@article{beguin_design_2003,
	series = {From {Computer} {Artefact} to {Instrument} for {Mediated} {Activity}.{Part} 1 {Organizational} {Issues}},
	title = {Design as a mutual learning process between users and designers},
	volume = {15},
	issn = {0953-5438},
	url = {https://www.sciencedirect.com/science/article/pii/S0953543803000602},
	doi = {10.1016/S0953-5438(03)00060-2},
	number = {5},
	urldate = {2026-03-20},
	journal = {Interacting with Computers},
	author = {Béguin, Pascal},
	month = oct,
	year = {2003},
	pages = {709--730},
}

@article{pintrich_conceptual_2004,
	title = {A {Conceptual} {Framework} for {Assessing} {Motivation} and {Self}-{Regulated} {Learning} in {College} {Students}},
	volume = {16},
	issn = {1573-336X},
	url = {https://doi.org/10.1007/s10648-004-0006-x},
	doi = {10.1007/s10648-004-0006-x},
	language = {en},
	number = {4},
	urldate = {2026-03-20},
	journal = {Educational Psychology Review},
	author = {Pintrich, Paul R.},
	month = dec,
	year = {2004},
	pages = {385--407},
}

@incollection{winne_studying_1998,
	address = {Mahwah, NJ, US},
	series = {The educational psychology series},
	title = {Studying as self-regulated learning},
	isbn = {978-0-8058-2481-0 978-0-8058-2482-7},
	booktitle = {Metacognition in educational theory and practice},
	publisher = {Lawrence Erlbaum Associates Publishers},
	author = {Winne, Philip H. and Hadwin, Allyson F.},
	year = {1998},
	pages = {277--304},
}

@article{boekaerts_self-regulated_1997,
	title = {Self-regulated learning: {A} new concept embraced by researchers, policy makers, educators, teachers, and students},
	volume = {7},
	issn = {0959-4752},
	shorttitle = {Self-regulated learning},
	url = {https://www.sciencedirect.com/science/article/pii/S0959475296000151},
	doi = {10.1016/S0959-4752(96)00015-1},
	number = {2},
	urldate = {2026-03-20},
	journal = {Learning and Instruction},
	author = {Boekaerts, Monique},
	month = jun,
	year = {1997},
	pages = {161--186},
}

@article{zimmerman_social_1989,
	address = {US},
	title = {A social cognitive view of self-regulated academic learning},
	volume = {81},
	issn = {1939-2176},
	doi = {10.1037/0022-0663.81.3.329},
	number = {3},
	journal = {Journal of Educational Psychology},
	publisher = {American Psychological Association},
	author = {Zimmerman, Barry J.},
	year = {1989},
	pages = {329--339},
}

@inproceedings{madaan_self-refine_2023,
	address = {Red Hook, NY, USA},
	series = {{NIPS} '23},
	title = {{SELF}-{REFINE}: iterative refinement with self-feedback},
	shorttitle = {{SELF}-{REFINE}},
	urldate = {2026-03-20},
	booktitle = {Proceedings of the 37th {International} {Conference} on {Neural} {Information} {Processing} {Systems}},
	publisher = {Curran Associates Inc.},
	author = {Madaan, Aman and Tandon, Niket and Gupta, Prakhar and Hallinan, Skyler and Gao, Luyu and Wiegreffe, Sarah and Alon, Uri and Dziri, Nouha and Prabhumoye, Shrimai and Yang, Yiming and Gupta, Shashank and Majumder, Bodhisattwa Prasad and Hermann, Katherine and Welleck, Sean and Yazdanbakhsh, Amir and Clark, Peter},
	month = dec,
	year = {2023},
	pages = {46534--46594},
}

@inproceedings{shinn_reflexion_2023,
	address = {Red Hook, NY, USA},
	series = {{NIPS} '23},
	title = {Reflexion: language agents with verbal reinforcement learning},
	shorttitle = {Reflexion},
	urldate = {2026-03-20},
	booktitle = {Proceedings of the 37th {International} {Conference} on {Neural} {Information} {Processing} {Systems}},
	publisher = {Curran Associates Inc.},
	author = {Shinn, Noah and Cassano, Federico and Gopinath, Ashwin and Narasimhan, Karthik and Yao, Shunyu},
	month = dec,
	year = {2023},
	pages = {8634--8652},
}

@inproceedings{panta_meda_2025,
	title = {{MEDA}: {A} {Multi}-{Agent} {System} {For} {Parametric} {CAD} {Model} {Creation}},
	shorttitle = {{MEDA}},
	url = {https://dx.doi.org/10.1115/DETC2025-163946},
	doi = {10.1115/DETC2025-163946},
	language = {en},
	urldate = {2026-03-20},
	publisher = {American Society of Mechanical Engineers Digital Collection},
	author = {Panta, Nirmal and Kafley, Saugat and Acharya, Rujal and Parajuli, Sashank and Parajuli, Dikshya and Panta, Prince and Belbase, Saroj and Pant, Sudikshya and Regmi, Amit and Tanaka, Akio and McComb, Christopher},
	month = oct,
	year = {2025},
}

@inproceedings{elrefaie_ai_2025,
	title = {{AI} {Agents} in {Engineering} {Design}: {A} {Multi}-{Agent} {Framework} for {Aesthetic} and {Aerodynamic} {Car} {Design}},
	shorttitle = {{AI} {Agents} in {Engineering} {Design}},
	url = {https://dx.doi.org/10.1115/DETC2025-169682},
	doi = {10.1115/DETC2025-169682},
	language = {en},
	urldate = {2026-03-20},
	publisher = {American Society of Mechanical Engineers Digital Collection},
	author = {Elrefaie, Mohamed and Qian, Janet and Wu, Raina and Chen, Qian and Dai, Angela and Ahmed, Faez},
	month = oct,
	year = {2025},
}

@article{massoudi_agentic_2026,
	title = {Agentic {Large} {Language} {Models} for {Conceptual} {Systems} {Engineering} and {Design1}},
	volume = {148},
	issn = {1050-0472},
	url = {https://doi.org/10.1115/1.4070328},
	doi = {10.1115/1.4070328},
	number = {051405},
	urldate = {2026-03-20},
	journal = {Journal of Mechanical Design},
	author = {Massoudi, Soheyl and Fuge, Mark},
	month = jan,
	year = {2026},
}

@incollection{salvi_how_2024,
	address = {Cambridge},
	title = {How {Impasse} {Leads} to {Insight}: {The} {Prepared} {Mind} {Perspective}},
	isbn = {978-1-009-24424-4},
	shorttitle = {How {Impasse} {Leads} to {Insight}},
	url = {https://www.cambridge.org/core/books/emergence-of-insight/how-impasse-leads-to-insight/0C4EE36632F265D2F463C732873AB32A},
	doi = {10.1017/9781009244244.005},
	urldate = {2026-03-20},
	booktitle = {The {Emergence} of {Insight}},
	publisher = {Cambridge University Press},
	author = {Seifert, Colleen M.},
	editor = {Salvi, Carola and Wiley, Jennifer and Smith, Steven M.},
	year = {2024},
	pages = {84--112},
}

@article{sio_fixation_2015,
	title = {Fixation or inspiration? {A} meta-analytic review of the role of examples on design processes},
	volume = {39},
	issn = {0142-694X},
	shorttitle = {Fixation or inspiration?},
	url = {https://www.sciencedirect.com/science/article/pii/S0142694X15000290},
	doi = {10.1016/j.destud.2015.04.004},
	urldate = {2026-03-20},
	journal = {Design Studies},
	author = {Sio, Ut Na and Kotovsky, Kenneth and Cagan, Jonathan},
	month = jul,
	year = {2015},
	pages = {70--99},
}

@article{yan_effects_2022,
	title = {Effects of self-assessment and peer-assessment interventions on academic performance: {A} meta-analysis},
	volume = {37},
	issn = {1747-938X},
	shorttitle = {Effects of self-assessment and peer-assessment interventions on academic performance},
	url = {https://www.sciencedirect.com/science/article/pii/S1747938X22000537},
	doi = {10.1016/j.edurev.2022.100484},
	urldate = {2026-03-20},
	journal = {Educational Research Review},
	author = {Yan, Zi and Lao, Hongling and Panadero, Ernesto and Fernández-Castilla, Belen and Yang, Lan and Yang, Min},
	month = nov,
	year = {2022},
	pages = {100484},
}

@article{sola-guirado_enhancing_2024,
	title = {Enhancing self-regulated learning in engineering education with lightboard videos as a support tool},
	volume = {32},
	copyright = {© 2024 The Authors. Computer Applications in Engineering Education published by Wiley Periodicals LLC.},
	issn = {1099-0542},
	url = {https://onlinelibrary.wiley.com/doi/abs/10.1002/cae.22756},
	doi = {10.1002/cae.22756},
	language = {en},
	number = {5},
	urldate = {2026-03-20},
	journal = {Computer Applications in Engineering Education},
	author = {Sola-Guirado, Rafael R. and Comino, Francisco and Castro-Triguero, Rafael},
	year = {2024},
	note = {\_eprint: https://onlinelibrary.wiley.com/doi/pdf/10.1002/cae.22756},
	pages = {e22756},
}

@article{xu_meta-analysis_2023,
	title = {A meta-analysis of the efficacy of self-regulated learning interventions on academic achievement in online and blended environments in {K}-12 and higher education},
	volume = {42},
	issn = {0144-929X},
	url = {https://doi.org/10.1080/0144929X.2022.2151935},
	doi = {10.1080/0144929X.2022.2151935},
	number = {16},
	urldate = {2026-03-20},
	journal = {Behaviour \& Information Technology},
	publisher = {Taylor \& Francis},
	author = {Xu, Zhihong and Zhao, Yingying and Zhang, Bingsheng and Liew, Jeffrey and Kogut, Ashlynn},
	month = dec,
	year = {2023},
	note = {\_eprint: https://doi.org/10.1080/0144929X.2022.2151935},
	pages = {2911--2931},
}

@article{jansen_self-regulated_2019,
	title = {Self-regulated learning partially mediates the effect of self-regulated learning interventions on achievement in higher education: {A} meta-analysis},
	volume = {28},
	issn = {1747-938X},
	shorttitle = {Self-regulated learning partially mediates the effect of self-regulated learning interventions on achievement in higher education},
	url = {https://www.sciencedirect.com/science/article/pii/S1747938X18304342},
	doi = {10.1016/j.edurev.2019.100292},
	urldate = {2026-03-20},
	journal = {Educational Research Review},
	author = {Jansen, Renée S. and van Leeuwen, Anouschka and Janssen, Jeroen and Jak, Suzanne and Kester, Liesbeth},
	month = nov,
	year = {2019},
	pages = {100292},
}

@article{lawanto_students_2010,
	title = {Students' metacognition during an engineering design project},
	volume = {23},
	issn = {1937-8327},
	url = {https://onlinelibrary.wiley.com/doi/abs/10.1002/piq.20084},
	doi = {10.1002/piq.20084},
	language = {en},
	number = {2},
	urldate = {2026-03-20},
	journal = {Performance Improvement Quarterly},
	author = {Lawanto, Oenardi},
	year = {2010},
	note = {\_eprint: https://onlinelibrary.wiley.com/doi/pdf/10.1002/piq.20084},
	pages = {117--136},
}

@inproceedings{lawanto_self-regulated_2014,
	title = {Self-regulated learning activities in engineering design education},
	issn = {2377-634X},
	url = {https://ieeexplore.ieee.org/document/7044475},
	doi = {10.1109/FIE.2014.7044475},
	urldate = {2026-03-20},
	booktitle = {2014 {IEEE} {Frontiers} in {Education} {Conference} ({FIE}) {Proceedings}},
	author = {Lawanto, Oenardi},
	month = oct,
	year = {2014},
	note = {ISSN: 2377-634X},
	pages = {1--4},
}

@inproceedings{gmeiner_exploring_2023,
	address = {New York, NY, USA},
	series = {{CHI} '23},
	title = {Exploring {Challenges} and {Opportunities} to {Support} {Designers} in {Learning} to {Co}-create with {AI}-based {Manufacturing} {Design} {Tools}},
	isbn = {978-1-4503-9421-5},
	url = {https://doi.org/10.1145/3544548.3580999},
	doi = {10.1145/3544548.3580999},
	booktitle = {Proceedings of the 2023 {CHI} {Conference} on {Human} {Factors} in {Computing} {Systems}},
	publisher = {Association for Computing Machinery},
	author = {Gmeiner, Frederic and Yang, Humphrey and Yao, Lining and Holstein, Kenneth and Martelaro, Nikolas},
	year = {2023},
}

@article{zheng_profiling_2020,
	title = {Profiling self-regulation behaviors in {STEM} learning of engineering design},
	volume = {143},
	issn = {0360-1315},
	url = {https://www.sciencedirect.com/science/article/pii/S0360131519302222},
	doi = {10.1016/j.compedu.2019.103669},
	urldate = {2026-03-20},
	journal = {Computers \& Education},
	author = {Zheng, Juan and Xing, Wanli and Zhu, Gaoxia and Chen, Guanhua and Zhao, Henglv and Xie, Charles},
	month = jan,
	year = {2020},
	pages = {103669},
}

@inproceedings{gmeiner_exploring_2025,
	address = {New York, NY, USA},
	series = {{DIS} '25},
	title = {Exploring the {Potential} of {Metacognitive} {Support} {Agents} for {Human}-{AI} {Co}-{Creation}},
	isbn = {979-8-4007-1485-6},
	url = {https://dl.acm.org/doi/10.1145/3715336.3735785},
	doi = {10.1145/3715336.3735785},
	urldate = {2026-03-20},
	booktitle = {Proceedings of the 2025 {ACM} {Designing} {Interactive} {Systems} {Conference}},
	publisher = {Association for Computing Machinery},
	author = {Gmeiner, Frederic and Luo, Kaitao and Wang, Ye and Holstein, Kenneth and Martelaro, Nikolas},
	month = jul,
	year = {2025},
	pages = {1244--1269},
}

@book{minsky_society_1986,
	address = {USA},
	title = {The society of mind},
	isbn = {0-671-60740-5},
	publisher = {Simon \& Schuster, Inc.},
	author = {Minsky, Marvin},
	year = {1986},
}

@article{liu_lost_2024,
	title = {Lost in the {Middle}: {How} {Language} {Models} {Use} {Long} {Contexts}},
	volume = {12},
	issn = {2307-387X},
	shorttitle = {Lost in the {Middle}},
	url = {https://doi.org/10.1162/tacl_a_00638},
	doi = {10.1162/tacl_a_00638},
	urldate = {2026-03-19},
	journal = {Transactions of the Association for Computational Linguistics},
	author = {Liu, Nelson F. and Lin, Kevin and Hewitt, John and Paranjape, Ashwin and Bevilacqua, Michele and Petroni, Fabio and Liang, Percy},
	month = feb,
	year = {2024},
	pages = {157--173},
}

@misc{zhang_recursive_2026,
	title = {Recursive {Language} {Models}},
	url = {http://arxiv.org/abs/2512.24601},
	doi = {10.48550/arXiv.2512.24601},
	urldate = {2026-03-19},
	publisher = {arXiv},
	author = {Zhang, Alex L. and Kraska, Tim and Khattab, Omar},
	month = jan,
	year = {2026},
	note = {arXiv:2512.24601 [cs]},
}




\end{document}